\documentclass[acmtog,authorversion]{acmart}
\acmSubmissionID{625}

\usepackage{amsmath}
\usepackage{bbm}
\usepackage{booktabs} 
\usepackage{enumitem}
\usepackage{epsfig}
\usepackage{graphicx}
\usepackage{makecell}
\usepackage{multirow}
\usepackage{subcaption}
\usepackage{xcolor}

\usepackage[ruled]{algorithm2e} 

\SetAlFnt{\small}
\SetAlCapFnt{\small}
\SetAlCapNameFnt{\small}
\SetAlCapHSkip{0pt}

\acmJournal{TOG}

\copyrightyear{2023}
\acmYear{2023}
\setcopyright{acmlicensed}
\acmConference[SIGGRAPH '23 Conference Proceedings]{Special Interest Group on Computer Graphics and Interactive Techniques Conference Conference Proceedings}{August 6--10, 2023}{Los Angeles, CA, USA}
\acmBooktitle{Special Interest Group on Computer Graphics and Interactive Techniques Conference Conference Proceedings (SIGGRAPH '23 Conference Proceedings), August 6--10, 2023, Los Angeles, CA, USA}
\acmPrice{15.00}
\acmDOI{10.1145/3588432.3591539}
\acmISBN{979-8-4007-0159-7/23/08}

\usepackage[capitalize]{cleveref}
\usepackage{textcomp}
\usepackage{amsfonts}
\crefname{section}{Sec.}{Secs.}
\Crefname{section}{Section}{Sections}
\Crefname{table}{Table}{Tables}
\crefname{table}{Tab.}{Tabs.}

\usepackage{scalerel}

\def\ie{i.e. }

\begin{document}
    \title{ViP-NeRF: Visibility Prior for Sparse Input Neural Radiance Fields}

\author{Nagabhushan Somraj}
\orcid{0000-0002-2266-759X}
\affiliation{%
  \institution{Indian Institute of Science}
  \city{Bengaluru}
  \country{India}
}
\email{nagabhushans@iisc.ac.in}
\author{Rajiv Soundararajan}
\affiliation{%
  \institution{Indian Institute of Science}
  \city{Bengaluru}
  \country{India}
}
\email{rajivs@iisc.ac.in}


    \begin{abstract}
        Neural radiance fields (NeRF) have achieved impressive performances in view synthesis by encoding neural representations of a scene.
        However, NeRFs require hundreds of images per scene to synthesize photo-realistic novel views.
        Training them on sparse input views leads to overfitting and incorrect scene depth estimation resulting in artifacts in the rendered novel views.
        Sparse input NeRFs were recently regularized by providing dense depth estimated from pre-trained networks as supervision, to achieve improved performance over sparse depth constraints.
        However, we find that such depth priors may be inaccurate due to generalization issues. 
        Instead, we hypothesize that the visibility of pixels in different input views can be more reliably estimated to provide dense supervision.
        In this regard, we compute a visibility prior through the use of plane sweep volumes, which does not require any pre-training.
        By regularizing the NeRF training with the visibility prior, we successfully train the NeRF with few input views.
        We reformulate the NeRF to also directly output the visibility of a 3D point from a given viewpoint to reduce the training time with the visibility constraint.
        On multiple datasets, our model outperforms the competing sparse input NeRF models including those that use learned priors.
        The source code for our model can be found on our project page: \url{https://nagabhushansn95.github.io/publications/2023/ViP-NeRF.html}.
    \end{abstract}

%
%


%
%


    \maketitle

    \begin{figure*}
        \centering
        \includegraphics[width=\linewidth]{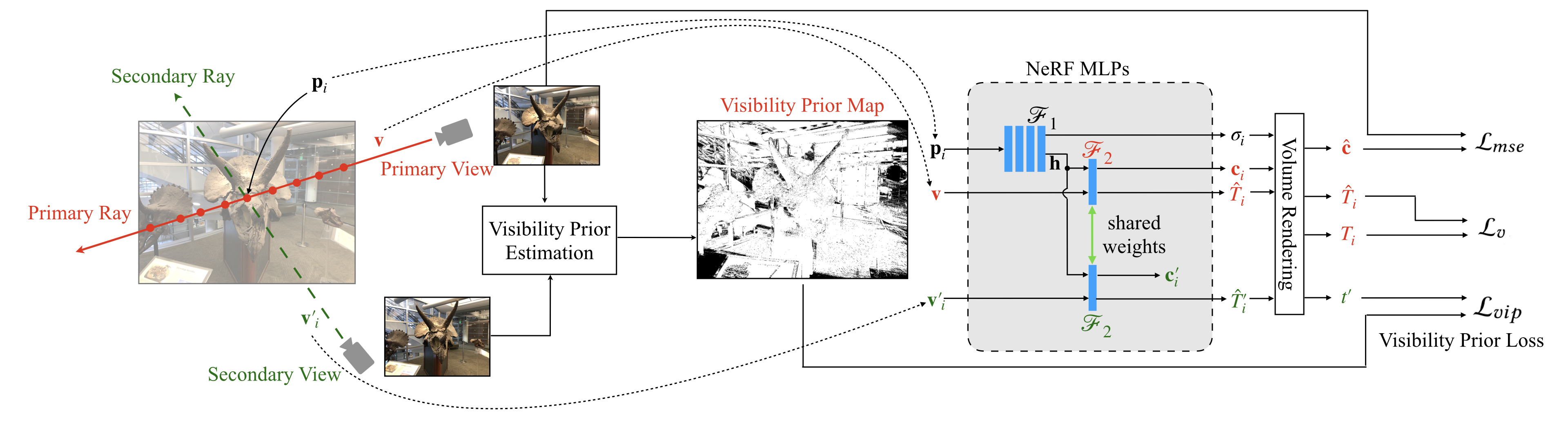}
        \caption{Overview of ViP-NeRF architecture.
        Given the images from \textcolor{red}{primary} and \textcolor[rgb]{0,0.45,0}{secondary} views, we estimate a visibility prior map in the primary view and use it to supervise the visibility of pixels as predicted by the NeRF.
        Specifically, we cast a ray through a randomly selected pixel in the primary view and sample 3D points along the ray.
        For every point $\mathbf{p}_i$, we use the NeRF MLPs to obtain its visibility in primary and secondary views, along with volume density $\sigma_i$ and color $\mathbf{c}_i$.
        Volume rendering outputs visibility $t'$ of the chosen pixel in the secondary view which is supervised by the visibility prior.
        $\mathcal{L}_{v}$ constrains the visibilities $\hat{T}_i$ output by network and $T_i$ computed using volume rendering to be consistent with each other.
        }
        \label{fig:model-architecture}
    \end{figure*}

    \section{Introduction}\label{sec:introduction}
    The goal of novel view synthesis is to synthesize a scene from novel viewpoints given RGB images of a few other viewpoints and their relative camera poses.
    By representing the scene implicitly using multi-layer perceptrons (MLP) and employing volume rendering, neural radiance fields (NeRF)~\cite{mildenhall2020nerf,barron2021mipnerf,barron2022mipnerf360,liu2022neuray} have achieved impressive view synthesis performance.
    Such superior performance is usually achieved when a large number of views is input to train the NeRF.
    However, in multiple applications such as virtual or augmented reality, telepresence, robotics, and autonomous driving, very few input images may be available for training~\cite{niemeyer2022regnerf}.
    In such settings, external sensors or a pre-calibrated fixed camera array may be employed to obtain accurate camera poses.
    Thus, there is a need to train NeRFs with few input views referred to as the sparse input NeRF problem.

    The key challenge with sparse input images is that the volume rendering equations in NeRF are under-constrained, leading to solutions that overfit the input views.
    This results in uncertain and inaccurate depth in the learned representation.
    Synthesized novel views in such cases contain extreme distortions such as blur, ghosting, and floater artifacts~\cite{niemeyer2022regnerf,roessle2022ddpnerf}.
    Recent works have proposed different approaches to constrain the training of NeRF to output visually pleasing novel views.
    While a few recent works~\cite{zhang2021ners,yang2022fvor,zhou2022sparsefusion} focus on training NeRF models on a specific category of objects such as chairs or airplanes, we focus on training category agnostic sparse input NeRF models~\cite{niemeyer2022regnerf}.
    Such prior work can be broadly classified into conditional NeRF models and other regularization approaches.

    The conditional NeRF models employ a latent representation of the scene obtained by pre-training on a large dataset of different scenes~\cite{yu2021pixelnerf,chen2021mvsnerf,wang2021ibrnet,hamdi2022sparf,johari2022geonerf} to condition the NeRF\@.
    The latent prior helps overcome the limitation on the number of views by enabling the NeRF model to effectively understand the scene.
    Such an approach is popular even when only a single image of the scene is available as input to the NeRF~\cite{xu2022sinnerf,lin2023vision,cai2022pix2nerf}.
    Different from the above, MetaNeRF~\cite{tancik2021metanerf} learns the latent information as initial weights of the NeRF MLPs by employing meta-learning.
    However, the pre-trained latent prior could suffer from poor generalization on a given target scene~\cite{niemeyer2022regnerf}.
    Thus, we believe that there is a need to study the sparse-input NeRF without conditioning the NeRF on latent representations.

    On the other hand, regularization based approaches constrain
    the NeRF training with novel loss functions to yield better solutions.
    DS-NeRF~\cite{deng2022dsnerf} uses sparse depth provided by a structure from motion (SfM) model as additional supervision for the NeRF. 
    To provide richer dense supervision, DDP-NeRF~\cite{roessle2022ddpnerf} completes the sparse depth map using a pre-trained convolutional neural network (CNN).
    However, the requirement of pre-training on a large dataset of scenes is cumbersome and the dense depth prior may suffer from generalization errors.
    RegNeRF~\cite{niemeyer2022regnerf} and InfoNeRF~\cite{kim2022infonerf} impose constraints to promote depth smoothness and reduce depth uncertainty respectively.
    However, in our experiments, we observe that these methods are still inferior to DS-NeRF on popular datasets.
    This motivates the exploration of other reliable features for dense supervision to constrain the NeRF in addition to sparse depth supervision.

    In our work, we explore the use of regularization in terms of visibility of any pixel from a pair of viewpoints.
    Here visibility of a pixel refers to whether the corresponding object is seen in both the viewpoints.
    For example, foreground objects are typically visible in multiple views whereas the background objects may be partially occluded.
    The visibility of a pixel in different views relies more on the relative depth of the scene objects than the absolute depth.
    We hypothesize that, given sparse input views, it may be easier to estimate the relative depth and visibility instead of the absolute depth.
    Thus, the key idea of our work is to regularize the NeRF with a dense visibility prior estimated using the given sparse input views.
    This allows the NeRF to learn better scene representation.
    We refer to our Visibility Prior regularized NeRF model as ViP-NeRF\@.

    To obtain the visibility prior, we employ the plane sweep volumes (PSV)~\cite{collins1996spacesweep} that have successfully been used in depth estimation~\cite{yang2003multiresolution,gallup2007realtime,ha2016highquality,im2019dpsnet} and view synthesis models~\cite{zhou2018stereomag}.
    We create the PSV by warping one of the images to the view of the other at different depths (or planes) and compare them to obtain error maps.
    We determine a binary visibility map for each pixel based on the corresponding errors in the PSV\@.
    We regularize the NeRF training by using such a map as supervision for every pair of input views.
    We use the visibility prior in conjunction with the depth prior from DS-NeRF~\cite{deng2022dsnerf}, where the former provides a dense prior on relative depth while the latter provides a sparse prior on absolute depth.
    Note that the estimation of our visibility prior does not require any pre-training on a large dataset.

    Regularizing the NeRF with a dense visibility prior is computationally intensive and can lead to impractical training times.
    We reformulate the NeRF to directly and additionally output visibility to impose the regularization in a computationally efficient manner.
    We conduct experiments on two popular datasets to demonstrate the efficacy of the visibility prior for sparse input NeRF\@.

    The main contributions of our work are as follows.
    \begin{itemize}
        \item We introduce visibility regularization to train the NeRF with sparse input views and refer to our model as ViP-NeRF.
        \item We estimate the dense visibility prior reliably using plane sweep volumes.
        \item We reformulate the NeRF MLP to output visibility thereby significantly reducing the training time.
        \item We achieve the state-of-the-art performance of sparse input NeRFs on multiple datasets.
    \end{itemize}

    \section{Related Work}\label{sec:related-work}
    \textbf{Novel View Synthesis}:
    Novel view synthesis methods typically use one or more input views to synthesize the scene from novel viewpoints.
    Recent pieces of work focus on obtaining volumetric 3D representations of the scene that can be computed once to render any viewpoint later.
    \citet{zhou2018stereomag} propose multi-plane image (MPI) representations for view synthesis.
    \citet{srinivasan2019pushing} further extend this by infilling the occluded regions in the MPIs.
    \citet{wiles2020synsin} study an extreme case with a single input image and generate novel views by employing a monocular depth estimation network for scene reprojection.
    In contrast to the above explicit representations, neural radiance fields~\cite{mildenhall2020nerf} use an implicit representation through coordinate-based neural networks.
    Although NeRFs achieve excellent performance, they require dense input views for training.
    In this work, we focus on solving this problem, i.e.\ to train a NeRF given very few input views.

    \textbf{Sparse Input NeRF}:
    Several recent works have studied sparse input NeRF by regularizing the NeRF with various priors.
    One of the early works, DietNeRF~\cite{jain2021dietnerf}, hallucinates novel viewpoints during training and constrains the NeRF to generate novel views similar to the input images in the CLIP~\cite{radford2021clip} representation space.
    DS-NeRF improves the performance by using fine-grained supervision at the pixel level using a sparse depth estimated by an SfM model~\cite{deng2022dsnerf}.
    DDP-NeRF further completes the sparse depth using a pre-trained network to obtain dense depth along with uncertainty estimates.
    Uncertainty modeling allows DDP-NeRF to relax the depth supervision at locations where the dense depth estimation is not confident.
    However, the completed depth may contain errors that may adversely affect the performance.
    DiffusioNeRF~\cite{wynn2023diffusionerf} instead employs a pre-trained denoising diffusion model to regularize the distribution of RGB-D patches in novel viewpoints.
    In contrast, our work uses a more reliable visibility prior which can be estimated without the use of sophisticated CNNs and does not require pre-training on a large dataset of scenes.

    Instead of depth estimates as priors, RegNeRF~\cite{niemeyer2022regnerf} regularizes the NeRF using depth smoothness constraints on the rendered patches in the hallucinated viewpoints.
    Different from depth regularization models, InfoNeRF~\cite{kim2022infonerf} tries to circumvent overfitting by encouraging concentration of volume density along a ray.
    In addition, it also minimizes the variation of volume density distributions along rays of two nearby viewpoints.
    Although these constraints are meaningful, our visibility prior imposes constraints across multiple views and can exploit the structure of the problem more effectively.

    \textbf{Single Image NeRF}:
    Recently, there is increased interest in training NeRFs with a single input image~\cite{xu2022sinnerf,lin2023vision}.
    A common thread in single image NeRF models is to use an encoder to obtain a latent representation of the input image.
    A NeRF based decoder conditioned on the representation, outputs volume density and color at given 3D points. 
    For example, pix2NeRF~\cite{cai2022pix2nerf} combines $\pi$-GAN~\cite{chan2021pigan} with NeRF to render photo-realistic images of objects or human faces.
    \citet{gao2020portrait} focus on human faces alone and use a more structured approach by exploiting facial geometry.
    MINE~\cite{li2021mine} combines NeRF with MPI by replacing the MLP based implicit representation with an MPI based explicit representation in the decoder.
    \citet{lin2023vision} obtain a richer latent representation by fusing global and local features obtained using a vision transformer and CNN respectively.
    Different from the above models, \citet{wimbauer2023behind} use the MLP decoder to predict volume density alone and obtain the color by directly sampling from the given images.
    However, a common drawback of these models is the need for pre-training.
    Thus the performance may be inferior when testing on a generic scene.

    \begin{figure}
        \centering
        \includegraphics[width=\linewidth]{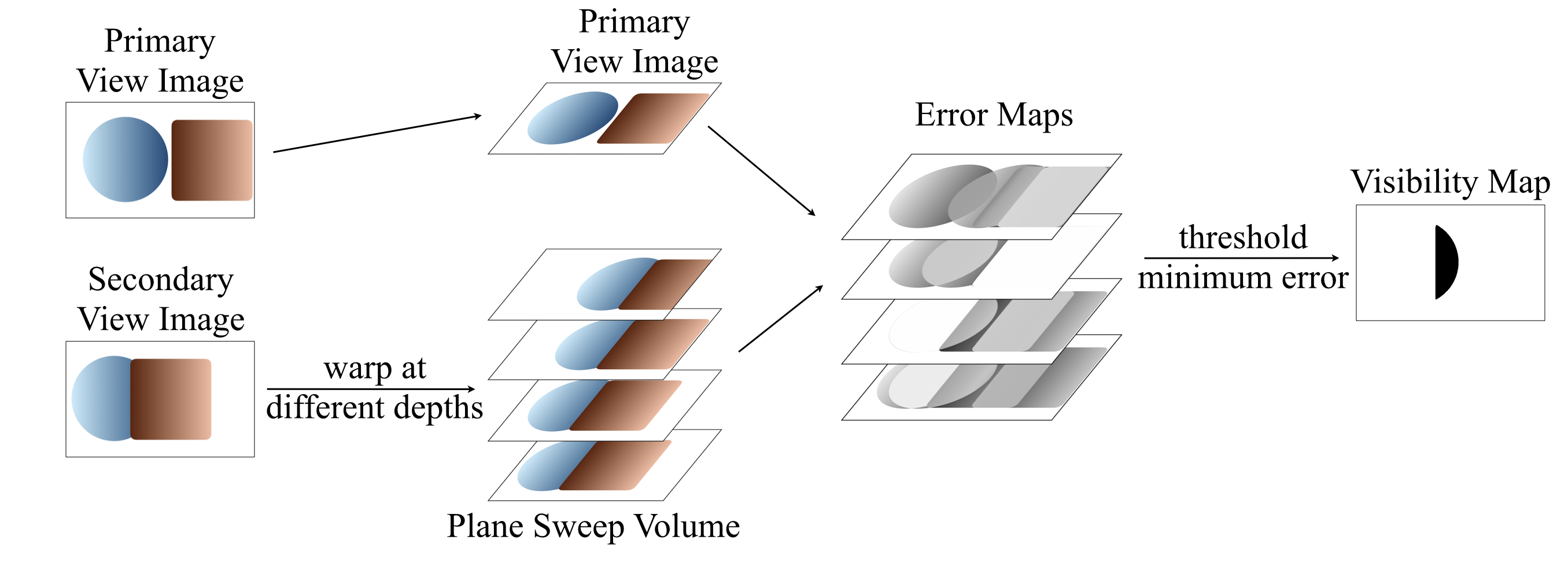}
        \caption{A toy example to illustrate the computation of visibility prior.
        The scene contains a blue sphere and a brown box and the relative pose between the views is a translation in $\mathbf{x}$ direction.
        The secondary view image is warped to the primary view at different depth planes to create a PSV and compared with the primary view image to obtain error maps.
        We observe that the brown square and the blue circle are matched better in the second and third planes respectively leading to lower error (denoted as white) in the respective error maps.
        The minimum error across all the planes is thresholded to obtain the visibility prior map corresponding to the primary view image.
        The right portion of the sphere which is occluded in the secondary view image is denoted in black in the visibility map.}
        \label{fig:visibility-prior-computation}
    \vspace{-5mm}
    \end{figure}


    \section{NeRF Preliminaries}\label{sec:nerf-preliminaries}
    We first provide a brief introduction to NeRF and define the notations for subsequent use.
    A neural radiance field is an implicit representation of a scene using two multi-layer perceptrons (MLP).
    Given a set of images of a scene with corresponding camera poses, a pixel $\mathbf{q}$ is selected at random, and a ray $\mathbf{r}$ is passed from the camera center $\mathbf{o}$ through $\mathbf{q}$.
    Let $\mathbf{p}_1, \mathbf{p}_2, \ldots, \mathbf{p}_N$ be $N$ randomly sampled 3D points along $\mathbf{r}$.
    If $\mathbf{d}$ is the direction vector of $\mathbf{r}$ and $z_i$ is the depth of a 3D point $\mathbf{p}_i$, $i \in \{ 1, 2, \ldots, N \}$, then $\mathbf{p}_i = \mathbf{o} + z_i \mathbf{d}$.
    An MLP $\mathcal{F}_1$ is trained to predict the volume density $\sigma_i$ at $\mathbf{p}_i$ as 
    \begin{align}
        \sigma_i, \mathbf{h}_i &= \mathcal{F}_1(\mathbf{p}_i), \label{eq:nerf-mlp1}
    \end{align}
    where $\mathbf{h}_i$ is a latent representation.
    A second MLP $\mathcal{F}_2$ then predicts the color using $\mathbf{h}_i$ and the viewing direction $\mathbf{v} = \mathbf{d} / \Vert \mathbf{d} \Vert$ as
    \begin{align}
        \mathbf{c}_i &= \mathcal{F}_2(\mathbf{h}_i, \mathbf{v}). \label{eq:nerf-mlp2}
    \end{align}
    Let the distance between two consecutive samples $\mathbf{p}_i$ and $\mathbf{p}_{i+1}$ be
    $ \delta_i = z_{i+1} - z_{i}$.
    The visibility or transmittance of $\mathbf{p}_i$ is then given by
    \begin{align}
        T_i &= \exp \left( - \sum_{j=1}^{i-1} \delta_j \sigma_j \right) \label{eq:volume-rendering-visibility}.
    \end{align}
    The weight or contribution of $\mathbf{p}_i$ in rendering the color $\hat{\mathbf{c}}$ of pixel $\mathbf{q}$ is computed as
     \begin{align}
         w_i = T_i \left( 1 - \exp(-\delta_i \sigma_i) \right) \label{eq:volume-rendering-weight}
     \end{align}
    to obtain
    \begin{align}
        \mathbf{\hat{c}} &= \sum_{i=1}^{N} w_i \mathbf{c}_i. \label{eq:volume-rendering-color}
    \end{align}
    The MLPs are trained using mean squared error loss with the true color $\mathbf{c}$ of $\mathbf{q}$ as
    \begin{align}
        \mathcal{L}_{mse} = \Vert \mathbf{c} - \mathbf{\hat{c}} \Vert^2. \label{eq:nerf-mse-loss}
    \end{align}

    \begin{table*}
        \centering
        \caption{Quantitative results on RealEstate-10K dataset.}
        \resizebox{\linewidth}{!}{
            \begin{tabular}{l|c|ccc|ccc|ccc}
                \hline
                & \textbf{learned} & \multicolumn{3}{c|}{2 views} & \multicolumn{3}{c|}{3 views} & \multicolumn{3}{c}{4 views} \\
                \textbf{Model} & \textbf{prior} &
                \textbf{LPIPS \textdownarrow} & \textbf{SSIM \textuparrow} & \textbf{PSNR \textuparrow} &
                \textbf{LPIPS \textdownarrow} & \textbf{SSIM \textuparrow} & \textbf{PSNR \textuparrow} &
                \textbf{LPIPS \textdownarrow} & \textbf{SSIM \textuparrow} & \textbf{PSNR \textuparrow} \\
                \hline
                InfoNeRF    &            & 0.6796          & 0.4653          & 12.30          & 0.6979          & 0.4024          & 11.15          & 0.6745          & 0.4298          & 11.52          \\
                DietNeRF    & \checkmark & 0.5730          & 0.6131          & 15.90          & 0.5365          & 0.6190          & 16.60          & 0.5337          & 0.6282          & 16.89          \\
                RegNeRF     &            & 0.5307          & 0.5709          & 16.14          & 0.4675          & 0.6096          & 17.38          & 0.4831          & 0.6068          & 17.46          \\
                DS-NeRF     &            & 0.4273          & 0.7223          & 21.40          & 0.3930          & 0.7554          & 23.73          & 0.3961          & 0.7575          & 24.24          \\
                DDP-NeRF    & \checkmark & 0.2527          & 0.7890          & 21.44          & 0.2240          & 0.8223          & 23.10          & 0.2190          & 0.8270          & 24.17          \\
                ViP-NeRF    &            & \textbf{0.1704} & \textbf{0.8087} & \textbf{24.48} & \textbf{0.1441} & \textbf{0.8505} & \textbf{27.21} & \textbf{0.1386} & \textbf{0.8588} & \textbf{28.13} \\  
                \hline
            \end{tabular}
        }
        \label{tab:quantitative-realestate}
    \end{table*}

    \begin{table*}
        \centering
        \caption{Quantitative results on NeRF-LLFF dataset.}
        \resizebox{\linewidth}{!}{
            \begin{tabular}{l|c|ccc|ccc|ccc}
                \hline
                & \textbf{learned} & \multicolumn{3}{c|}{2 views} & \multicolumn{3}{c|}{3 views} & \multicolumn{3}{c}{4 views} \\
                \textbf{Model} & \textbf{prior} &
                \textbf{LPIPS \textdownarrow} & \textbf{SSIM \textuparrow} & \textbf{PSNR \textuparrow} &
                \textbf{LPIPS \textdownarrow} & \textbf{SSIM \textuparrow} & \textbf{PSNR \textuparrow} &
                \textbf{LPIPS \textdownarrow} & \textbf{SSIM \textuparrow} & \textbf{PSNR \textuparrow} \\
                \hline
                InfoNeRF          &            & 0.7561          & 0.2095          & 9.23           & 0.7679          & 0.1859          & 8.52           & 0.7701          & 0.2188          & 9.25           \\
                DietNeRF          & \checkmark & 0.7265          & 0.3209          & 11.89          & 0.7254          & 0.3297          & 11.77          & 0.7396          & 0.3404          & 11.84          \\
                RegNeRF           &            & 0.4402          & 0.4872          & 16.90          & 0.3800          & 0.5600          & 18.62          & \textbf{0.3446} & 0.6056          & 19.83          \\
                DS-NeRF           &            & 0.4548          & 0.5068          & 17.06          & 0.4077          & 0.5686          & \textbf{19.02} & 0.3825          & 0.6016          & \textbf{20.11} \\
                DDP-NeRF          & \checkmark & 0.4223          & \textbf{0.5377} & \textbf{17.21} & 0.4178          & 0.5610          & 17.90          & 0.3821 & 0.5999 & 19.19 \\
                ViP-NeRF          &            & \textbf{0.4017} & 0.5222          & 16.76          & \textbf{0.3750} & \textbf{0.5837} & 18.92          & 0.3593 & \textbf{0.6085} & 19.57      \\  
                \hline
            \end{tabular}
        }
        \label{tab:quantitative-llff}
    \end{table*}

    \section{Method}\label{sec:method}
    We illustrate the outline of our model in \cref{fig:model-architecture}.
    The core idea of our work is that when only a few multiview images are available for NeRF training, the visibility of a pixel in different views can be more reliably densely estimated as compared to its absolute depth.
    In this regard, we introduce visibility regularization to train the NeRF with sparse input views in \cref{subsec:visibility-regularization}.
    To impose the visibility regularization, we obtain a binary visibility prior map for every pair of input training images, which we explain in \cref{subsec:visibility-prior}.
    Finally, to reduce the training time, we design a method to efficiently predict the visibility of a given pixel in different views in \cref{subsec:efficient-visibility-prediction}.
    \cref{subsec:overall-loss} summarizes the various loss functions used in training our model.

    \subsection{Visibility Regularization}\label{subsec:visibility-regularization}
    Recall from \cref{sec:nerf-preliminaries} that NeRF trains MLPs by picking a random pixel $\mathbf{q}$ and predicting the color of $\mathbf{q}$ using the MLPs and volume rendering.
    Without loss of generality, we refer to the view corresponding to the ray $\mathbf{r}$ passing through $\mathbf{q}$ as the primary view and choose any other view as a secondary view.
    NeRF then samples $N$ candidate 3D points, $\mathbf{p}_1, \mathbf{p}_2, \ldots, \mathbf{p}_N$, along $\mathbf{r}$.
    Let $T'_i$ be the visibility of $\mathbf{p}_i$ from the secondary view, computed similar to \cref{eq:volume-rendering-visibility}.
    We define the visibility of pixel $\mathbf{q}$ in the secondary view, $t'(\mathbf{q})$, as the weighted visibilities of all the candidate 3D points $\mathbf{p}_i$ analogous to \cref{eq:volume-rendering-color} as
    \begin{align}
        t'(\mathbf{q}) = \sum_{i=1}^{N} w_i T'_i \quad \in [0,1], \label{eq:volume-rendering-pixel-visibility2}
    \end{align}
    where $w_i$ are obtained through \cref{eq:volume-rendering-weight}.
    We omit the dependence of $w_i$ and $T'_i$ on $\mathbf{q}$ in the above equation for ease of reading.
    We obtain a prior $\tau'(\mathbf{q}) \in \{0, 1\}$ on the visibility $t'(\mathbf{q})$ as described in \cref{subsec:visibility-prior}.
    We constrain the visibility $t'(\mathbf{q})$ to match the prior $\tau'(\mathbf{q})$.
    However, we find that the prior may be unreliable at pixels where $\tau' = 0$, as we describe in \cref{subsec:visibility-prior}.
    Hence, we do not impose any visibility loss on such pixels and formulate our visibility prior loss as
    \begin{align}
        \mathcal{L}_{vip}(\mathbf{q}) = \max(\tau'(\mathbf{q}) - t'(\mathbf{q}), 0). \label{eq:visibility-prior-loss}
    \end{align}
    Note that our loss function constrains the NeRF across pairs of views, unlike previous works which regularize~\cite{roessle2022ddpnerf,niemeyer2022regnerf} in a given view alone.
    We believe that this leads to a better regularization for synthesizing novel views.


    \subsection{Visibility Prior}\label{subsec:visibility-prior}
    Given primary and secondary views, our goal is to estimate whether every pixel in the primary view is also visible in the secondary view through a binary visibility prior $\tau'(\mathbf{q})$.
%
    We employ plane sweep volumes to compute the visibility prior.
    We illustrate the computation of the visibility prior with a toy example in \cref{fig:visibility-prior-computation}.
    Here, we warp the image in the secondary view to the primary view using the camera parameters at different depths varying between the near depth $z_{\min}$ and far depth $z_{\max}$.
    We sample $D$ depths uniformly in inverse depth similar to StereoMag~\cite{zhou2018stereomag}.
    The set of warped images is referred to as plane sweep volume (PSV)~\cite{huang2018deepmvs}.

    Let $I^{(1)}$ be the image in the primary view and $I_k^{(2)}$ be the set of $D$ warped images, where $k \in \{0, 1, \ldots, D-1\}$ denotes the plane index.
    We then compute the error map $E_k$ of the warped secondary image with the primary image at each plane $k$ of the PSV as
    \begin{align}
        E_k = \Vert I^{(1)} - I_k^{(2)} \Vert_1, \label{eq:psv-error}
    \end{align}
    where the norm is computed across the color channels.
    We determine the visibility prior $\tau'$ for pixel $\mathbf{q}$ by thresholding the minimum error across all the planes as
    \begin{align}
        \nonumber e(\mathbf{q}) &= \min_k E_k(\mathbf{q}), \\
        \tau'(\mathbf{q}) &= \mathbbm{1}_{\{ \exp{(-e(\mathbf{q})/\gamma)} > 0.5 \} }, \label{eq:visibility-prior-estimation}
    \end{align}
    where $\gamma$ is a hyper-parameter.

    Intuitively, for a given pixel $\mathbf{q}$, a lower error in any of the planes indicates the presence of a matching pixel in the secondary view, \ie $\mathbf{q}$ is visible in the secondary view.
    Note that this holds true when the intensity of pixels does not change significantly across views, which is typical for most of the objects in real-world scenes~\cite{li2021mine}.
    Consequently, the absence of a matching point across all the planes may indicate that $\mathbf{q}$ is not visible in the secondary view or $\mathbf{q}$ belongs to a highly specular object whose color varies significantly across different viewpoints.
    Thus, our prior is used to regularize the NeRF only in the first case above \ie the pixels for which we find a match.
    Following the above procedure, we obtain the visibility prior for every pair of images obtained from the training set, by treating either image in the pair as the primary or the secondary view.

    \begin{figure*}
        \centering
        \includegraphics[width=\linewidth]{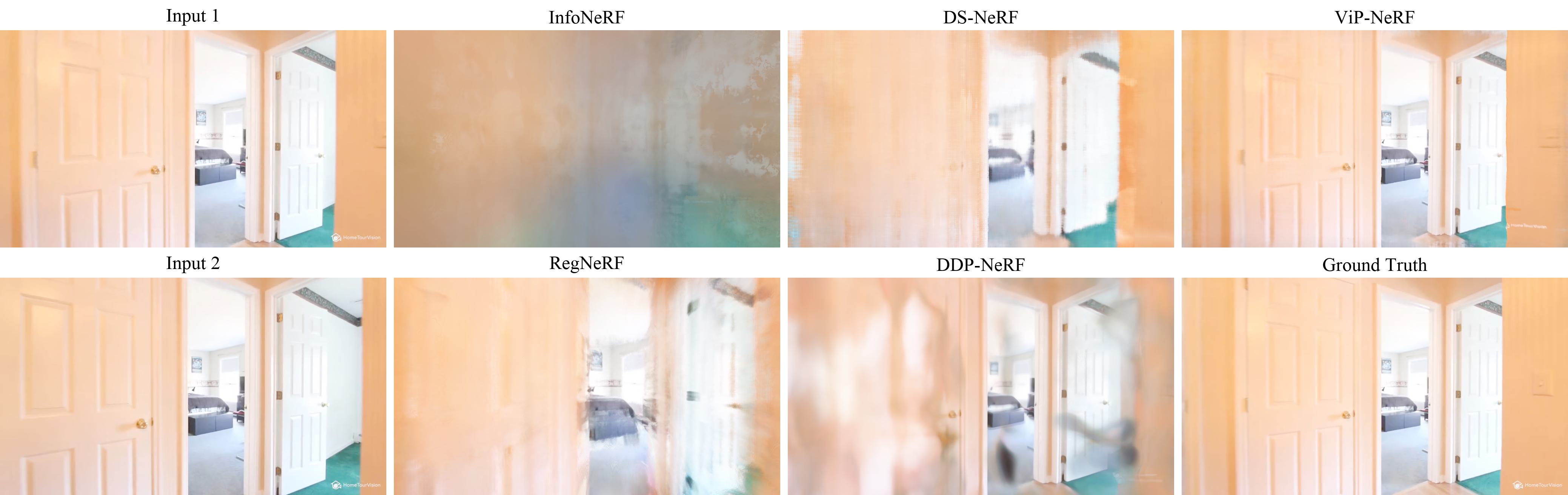}
        \caption{Qualitative examples on RealEstate-10K dataset with two input views.
        We observe that the predictions of ViP-NeRF are close to the ground truth, while those of other models suffer from various distortions.
        In particular, DDP-NeRF blurs regions of the frame near the left door and contains black floater artifacts.
        }
        \label{fig:qualitative-realestate01a}
    \end{figure*}

    \subsection{Efficient Prediction of Visibility}\label{subsec:efficient-visibility-prediction}
    Recall that imposing $\mathcal{L}_{vip}$ in \cref{eq:visibility-prior-loss} requires computing visibility $T'_i$ in the secondary view for every $\mathbf{p}_i$.
    A naive approach to compute $T'_i$ involves sampling up to $N$ points along a secondary ray from the secondary view camera origin to $\mathbf{p}_i$ and querying the NeRF MLP $\mathcal{F}_1$ for each of these points.
    Thus, obtaining $t'(\mathbf{q})$ in \cref{eq:volume-rendering-pixel-visibility2} requires upto $N^2$ MLP queries, which increases the training time making it computationally prohibitive.
    We overcome this limitation by reformulating the NeRF MLP $\mathcal{F}_2$ to also output a view-dependent visibility of a given 3D point as,
    \begin{align}
        \mathbf{c}_i, \hat{T}_i = \mathcal{F}_2(\mathbf{h}_i, \mathbf{v}); \quad
        \mathbf{c}'_i, \hat{T}'_i = \mathcal{F}_2(\mathbf{h}_i, \mathbf{v}'_i), \label{eq:visibility-prediction}
    \end{align}
    where $\mathbf{v}'_i$ is the viewing direction of the secondary ray.
    We use the MLP output $\hat{T}'_i$ instead of $T'_i$ in \cref{eq:volume-rendering-pixel-visibility2}.

    Note that to output $\hat{T}'_i$, we need not query $\mathcal{F}_1$ again and can reuse $\mathbf{h}_i$ obtained from \cref{eq:nerf-mlp1}.
    We only need to query $\mathcal{F}_2$ additionally and since $\mathcal{F}_2$ is a single layer MLP and significantly smaller than $\mathcal{F}_1$, the additional computational burden is negligible.
    Thus, directly obtaining the secondary visibility $\hat{T}'_i$ of $\mathbf{p}_i$ through \cref{eq:visibility-prediction} allows us to compute $t'(\mathbf{q})$ in \cref{eq:volume-rendering-pixel-visibility2} using only $N$ queries of the MLP $\mathcal{F}_1$, as opposed to $N^2$ queries in the naive approach.

    However, the use of $\hat{T}'_i$ in place of $T'_i$ regularizes the NeRF training only if the two quantities are close to each other.
    Thus, we introduce an additional loss to constrain the visibility $\hat{T}_i$ output by $\mathcal{F}_2$ to be consistent with the visibility $T_i$ computed using \cref{eq:volume-rendering-visibility} as
    \begin{align}
        \mathcal{L}_{v} = \sum_{i=1}^{N} \left( \left( \text{SG}(T_i) - \hat{T}_i \right)^2 + \left( T_i - \text{SG}(\hat{T}_i) \right)^2 \right), \label{eq:visibility-loss}
    \end{align}
    where $\text{SG}(\cdot)$ denotes the stop-gradient operation.
    The first term in the above loss function uses $T_i$ as a target and brings $\hat{T}_i$ closer to it.
    On the other hand, since $\hat{T}_i$ gets additionally updated directly based on the visibility prior, the second term helps transfer such updates to $\mathcal{F}_1$ more efficiently than backpropagation through $\mathcal{F}_2$.


    \subsection{Overall Loss}\label{subsec:overall-loss}
    Similar to DS-NeRF~\cite{deng2022dsnerf}, we also use the sparse depth given by an SfM model to supervise the NeRF as
    \begin{align}
        \mathcal{L}_{sd} = \Vert z - \hat{z} \Vert^2 , \label{eq:sparse-depth-loss}
    \end{align}
    where $z$ is the depth provided by the SfM model, $\hat{z} = \sum_i w_i z_i$ is the depth estimated by NeRF and $w_i$ are obtained in \cref{eq:volume-rendering-weight}.
    Our overall loss for ViP-NeRF is a linear combination of the losses obtained in \cref{eq:nerf-mse-loss}, \cref{eq:visibility-prior-loss}, \cref{eq:visibility-loss} and \cref{eq:sparse-depth-loss} as
    \begin{align}
        \mathcal{L} = \lambda_1 \mathcal{L}_{mse} + \lambda_2 \mathcal{L}_{sd} + \lambda_3 \mathcal{L}_{vip} + \lambda_4 \mathcal{L}_{v},
    \end{align}
    where $\lambda_1, \lambda_2, \lambda_3 \text{ and } \lambda_4$ are hyper-parameters.
    We note that $\mathcal{L}_{vip}$ is always employed in conjunction with $\mathcal{L}_{v}$ to make the learning computationally tractable.

    \section{Experiments}\label{sec:experiments}
    \begin{figure}
        \centering
        \includegraphics[width=\linewidth]{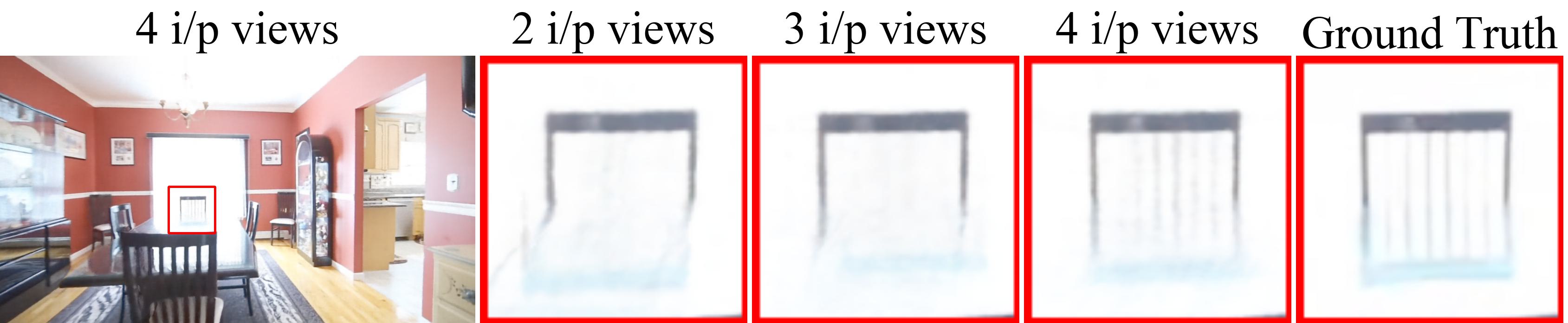}
        \includegraphics[width=\linewidth]{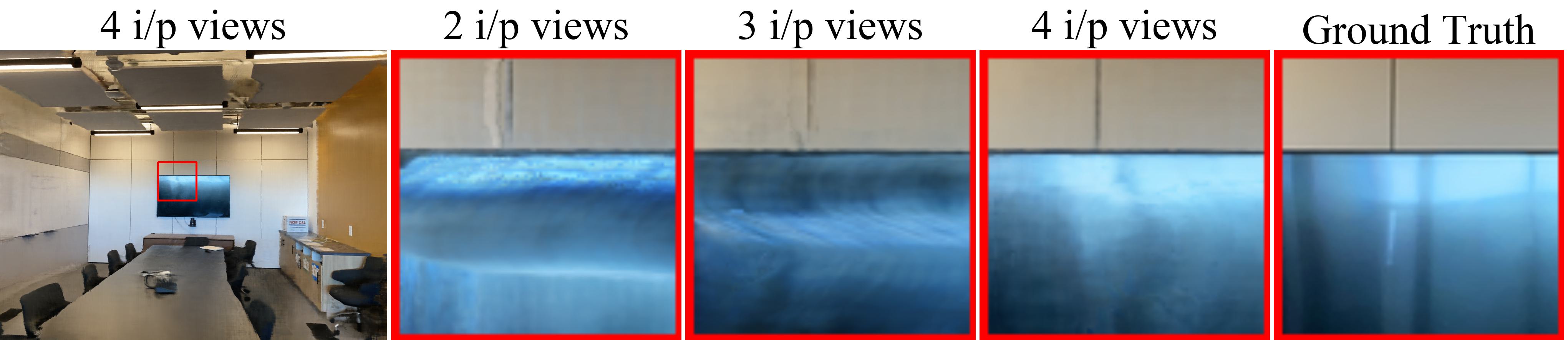}
        \caption{Qualitative examples on RealEstate-10K and NeRF-LLFF dataset with two, three, and four input views.
        We observe that ViP-NeRF models specular regions better as the number of input views increases.
        For example, in the first row, the reflection of the chair is better reconstructed as the number of views increases.}
        \label{fig:qualitative-progress}
    \end{figure}


    \subsection{Evaluation Setup}\label{subsec:evaluation-setup}
    We conduct experiments on two different datasets, namely RealEstate-10K and NeRF-LLFF\@.
    We evaluate all the models in the more challenging setup of 2, 3, or 4 input views, unlike prior work which use 9--18 input views~\cite{jain2021dietnerf,roessle2022ddpnerf}.
    The test set is retained to be the same across all different settings for both datasets.
    \vspace{1mm}

    \textbf{RealEstate-10K}~\cite{zhou2018stereomag} dataset is commonly used to evaluate view synthesis models~\cite{tucker2020single,han2022single} and contains videos of camera motion, both indoor and outdoor.
    The dataset also provides the camera intrinsics and extrinsics for all the frames.
    For our experiments, we choose 5 scenes from the test set, each containing 50 frames with a spatial resolution of $1024 \times 576$.
    In each scene, we reserve every $10^\text{th}$ frame for training and use the remaining 45 frames for testing.
    Please refer to the supplementary for more details on the choice of scenes.
    \vspace{1mm}


    \textbf{NeRF-LLFF}~\cite{mildenhall2019llff} dataset is used to evaluate the performance of various NeRF Models including sparse input NeRF models.
    It consists of 8 forward-facing scenes with a variable number of frames per scene at a spatial resolution of $1008 \times 756$.
    Following RegNeRF~\cite{niemeyer2022regnerf}, we use every $8^\text{th}$ frame for testing.
    For training, we pick 2, 3 or 4 frames uniformly among the remaining frames following RegNeRF~\cite{niemeyer2022regnerf}.
    \vspace{1mm}

    \textbf{Evaluation measures}:
    We quantitatively evaluate the methods using LPIPS~\cite{zhang2018unreasonable}, structural similarity (SSIM)~\cite{wang2004image}, and peak signal to noise ratio (PSNR) measures.
    For LPIPS, we use the v0.1 release with the AlexNet~\cite{krizhevsky2012alexnet} backbone as suggested by the authors.

    \begin{table}
        \centering
        \setlength\tabcolsep{4.5pt}
        \caption{Comparison of reliability of priors used in different models.
        The reference visibility is obtained using NeRF trained with dense input views.}
        \begin{tabular}{c|ccc|ccc}
            \hline
            & \multicolumn{3}{c|}{RealEstate-10K} & \multicolumn{3}{c}{NeRF-LLFF} \\
            model &
            \textbf{Prec. \textuparrow} & \textbf{Rec. \textuparrow} & \textbf{F1 \textuparrow} &
            \textbf{Prec. \textuparrow} & \textbf{Rec. \textuparrow} & \textbf{F1 \textuparrow} \\
            \hline
            ViP-NeRF & 0.97          & \textbf{0.83} & \textbf{0.89} & 0.82          & \textbf{0.85} & \textbf{0.83} \\
            DDP-NeRF & \textbf{0.98} & 0.53          & 0.66          & \textbf{0.86} & 0.33          & 0.47          \\
            \hline
        \end{tabular}
        \label{tab:quantitative-prior}
    \end{table}

    \begin{table}
        \centering
        \setlength\tabcolsep{4.5pt}
        \caption{Evaluation of depth estimated by different models with two input views.
        The reference depth is obtained using NeRF trained with dense input views.
        The depth RMSE on the two datasets are of different orders on account of different depth ranges.}
        \begin{tabular}{c|cc|cc}
            \hline
            & \multicolumn{2}{c|}{RealEstate-10K} & \multicolumn{2}{c}{NeRF-LLFF} \\
            model &
            \textbf{RMSE \textdownarrow} & \textbf{SROCC \textuparrow} &
            \textbf{RMSE \textdownarrow} & \textbf{SROCC \textuparrow} \\
            \hline
            ViP-NeRF & \textbf{1.6411} & \textbf{0.7702} & \textbf{45.6314} & \textbf{0.6184} \\
            DDP-NeRF & 1.7211 & 0.7544 & 46.6268 & 0.6136 \\
            \hline
        \end{tabular}
        \label{tab:quantitative-depth}
    \end{table}

    \subsection{Comparisons and Implementation Details}\label{subsec:comparisons}
    We compare the performance of our model with other sparse input NeRF models such as DDP-NeRF~\cite{roessle2022ddpnerf} and DietNeRF~\cite{jain2021dietnerf} which use learned priors to constrain the NeRF training.
    We also compare with DS-NeRF~\cite{deng2022dsnerf}, InfoNeRF~\cite{kim2022infonerf}, and RegNeRF~\cite{niemeyer2022regnerf} that do not use learned priors.
    We train the models for 50k iterations on both datasets using the code provided by the respective authors.

    For ViP-NeRF, we use Adam optimizer with a learning rate of 5e-4 that exponentially decays to 5e-6 following NeRF~\cite{mildenhall2020nerf}.
    We set the loss weights such that the magnitudes of all the losses are of similar order after scaling.
    Specifically, we set $\lambda_1 = 1, \lambda_2 = 0.1, \lambda_3 = 0.001 \text{ and } \lambda_4 = 0.1$.
    For visibility prior estimation, we set $D=64$ and $\gamma=10$.
    Since we require $\hat{T}'_i$ to be close to $T'_i$ while using $\hat{T}'_i$ to compute $\mathcal{L}_{vip}$, we impose $\mathcal{L}_{vip}$ after 20k iterations.
    We train our models on a single NVIDIA RTX A4000 16GB GPU\@.

    \subsection{Results}\label{subsec:results}
    We show the quantitative performance of ViP-NeRF and other competing models on RealEstate-10K and NeRF-LLFF datasets in \cref{tab:quantitative-realestate,tab:quantitative-llff}.
    Our model outperforms all the competing models, particularly in terms of the perceptual metric, LPIPS.
    ViP-NeRF even outperforms models such as DDP-NeRF and DietNeRF which involve pre-training on a large dataset.
    \cref{fig:qualitative-realestate01a} shows qualitative comparisons on a scene from the RealEstate-10K dataset, where we observe significantly better synthesis by our model as compared to the competing models.
    We show more qualitative comparisons in \cref{fig:qualitative-realestate01b,fig:qualitative-realestate02,fig:qualitative-realestate03,fig:qualitative-llff01} in the figure only pages at the end of this manuscript.
    In these samples, we find that ViP-NeRF removes most of the floater artifacts and successfully retains the shapes of objects.

    \begin{figure}
        \centering
        \includegraphics[width=\linewidth]{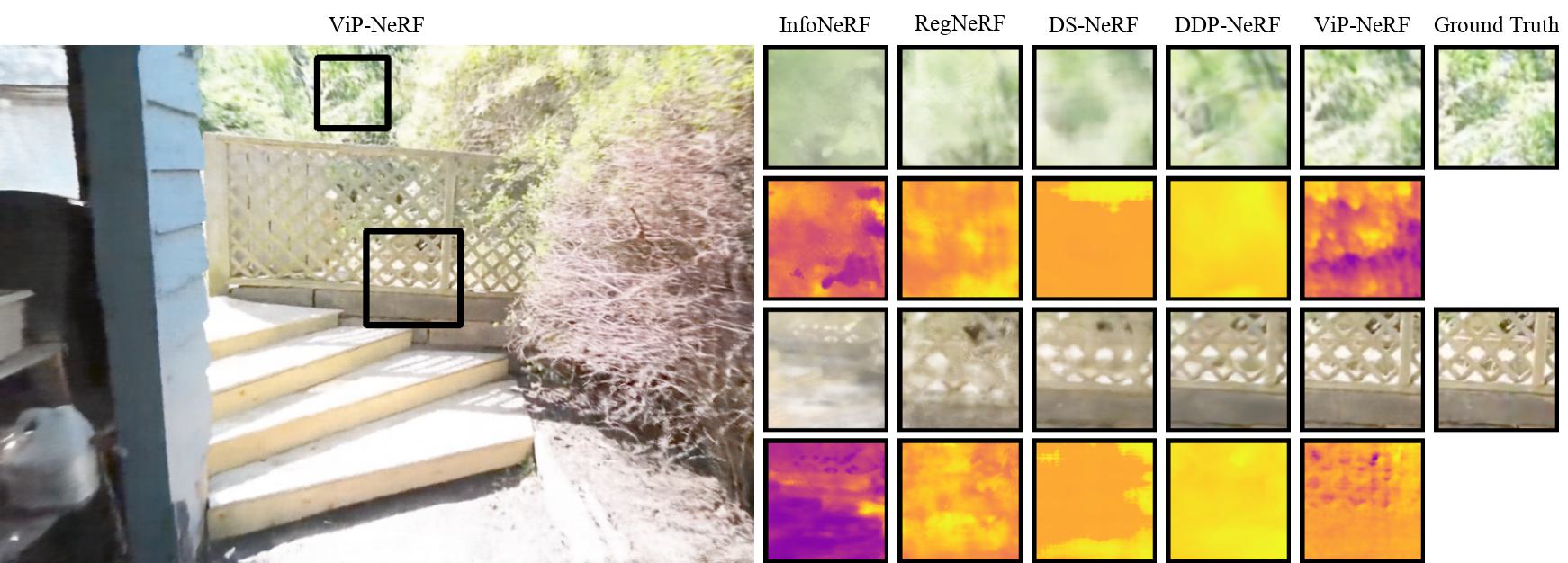}
        \caption{Estimated depth map on RealEstate-10K dataset with two input views.
        We find that ViP-NeRF is better in both frame synthesis and depth estimation compared to the competing models.
        For example, in the first row, the depth estimated by DDP-NeRF is smooth which may be leading to a loss of sharpness in synthesizing the shrubs.
        In contrast, ViP-NeRF predictions are sharper.
        For better visualization, we show inverse depth and normalize it to set the maximum value to unity.}
        \label{fig:qualitative-depth}
    \end{figure}

    \begin{figure}
        \centering
        \includegraphics[width=\linewidth]{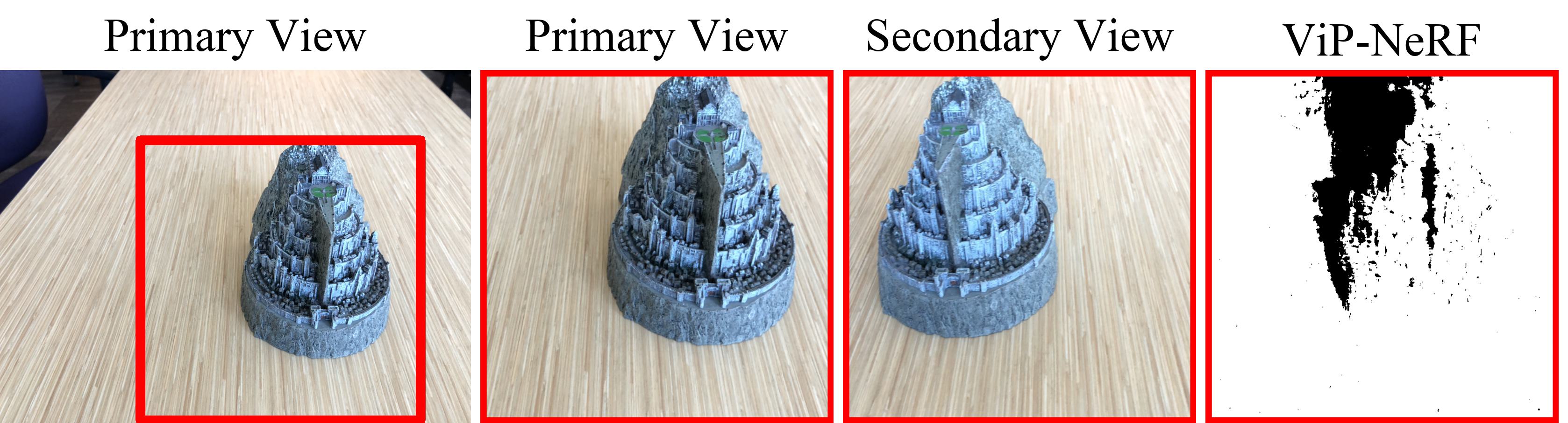}
        \caption{Visualization of the visibility map predicted by ViP-NeRF.
        White indicates the regions of the `Primary View' which are visible in the `Secondary View' and black indicates the occluded regions.
        From the primary and secondary views, we observe that the left part of the fortress and the neighboring portion of the wood are hidden in the secondary view.
        ViP-NeRF is able to reasonably determine the visible and occluded regions.}
        \label{fig:qualitative-visibility}
    \end{figure}


    \begin{table}
        \centering
        \caption{Ablation experiments on both the datasets with two input views.}
        \begin{tabular}{c|cc|cc}
            \hline
            & \multicolumn{2}{c|}{RealEstate-10K} & \multicolumn{2}{c}{NeRF-LLFF} \\
            model                & \textbf{LPIPS \textdownarrow} & \textbf{SSIM \textuparrow} & \textbf{LPIPS \textdownarrow} & \textbf{SSIM \textuparrow} \\
            \hline
            ViP-NeRF             & \textbf{0.1704}               & \textbf{0.8087}            & \textbf{0.4017}               & \textbf{0.5222}            \\  
            w/o sparse depth     & 0.2754                        & 0.7588                     & 0.5056                        & 0.4631                     \\  
            w/o dense visibility & 0.4273                        & 0.7223                     & 0.4548                        & 0.5068                     \\  
            \hline
        \end{tabular}
        \label{tab:quantitative-ablations01}
    \end{table}


    In \cref{fig:qualitative-progress}, we qualitatively compare the predictions of our model with different numbers of input views.
    We observe that ViP-NeRF estimates the geometry reasonably well with even two input views.
    However, with more input views, the performance of ViP-NeRF improves in reflective or specular regions.
    \cref{fig:qualitative-visibility} visualizes the visibility map predicted by ViP-NeRF, where we observe that it is able to accurately predict the regions in the primary image which are visible and occluded in the secondary image.
    \vspace{1mm}



    \textbf{Dense depth vs dense visibility:}
    The key idea of our paper is that it may be possible to reliably estimate dense visibility than dense depth.
    From \cref{tab:quantitative-realestate}, we find that ViP-NeRF outperforms DDP-NeRF consistently, which indicates that the dense visibility prior we compute without any pre-training is superior to the learned dense depth prior used by DDP-NeRF\@.
    Further from \cref{tab:quantitative-llff}, we observe that ViP-NeRF consistently improves over DS-NeRF in terms of LPIPS and SSIM, whereas DDP-NeRF does not.
    This may be due to the domain shift between the training dataset of DDP-NeRF and the LLFF dataset, resulting in no performance improvement over DS-NeRF\@.
    Thus, we conclude that augmenting sparse depth with dense visibility leads to better view synthesis performance than dense completion of the sparse depth.
    We further validate this conclusion by comparing the two priors in the following.
    \vspace{1mm}


    \textbf{Validating priors:}
    We compare the reliability of the dense visibility prior used in our model against the dense depth prior from DDP-NeRF\@.
    For this comparison, we convert the dense depth to visibility and compare it with the visibility prior of our approach.
    Specifically, we warp the image in the secondary view to the primary view using the dense depth prior and compute the visibility map similar to \cref{eq:visibility-prior-estimation}.
    We compare the visibility maps obtained using dense depth and our approach with the visibility map predicted by a NeRF model trained with dense input views.
    We evaluate the visibility maps in terms of precision, recall, and F1 score.

    From \cref{tab:quantitative-prior}, we observe that our approach significantly outperforms DDP-NeRF prior in terms of the recall and F1 score, while performing similarly in terms of precision.
    A high precision of our prior indicates that it makes very few mistakes when imposing $\mathcal{L}_{vip}$.
    On the other hand, a high recall shows that our prior is able to capture most of the visible regions where $\mathcal{L}_{vip}$ needs to be imposed.
    On the contrary, a low recall for the DDP-NeRF prior indicates that large regions that are actually visible in the secondary view are marked as occluded by the dense depth prior.
    Consequently, this indicates the presence of a large number of pixels with inaccurate depth in the prior of DDP-NeRF.
    Thus, we conclude that our visibility prior is more reliable than the dense depth prior from DDP-NeRF for training the NeRF.
    
    As discussed in \cref{sec:introduction}, visibility is related to relative depth, and thus a prior on visibility only constrains the relative depth ordering of the objects.
    On the other hand, the dense depth prior constrains the absolute depth, perhaps incorrectly.
    Thus the visibility prior provides more freedom to the NeRF in reconstructing the 3D geometry and is also more reliable compared to the depth prior.
    This may explain the superior performance of visibility regularization over dense depth regularization.
    \vspace{1mm}

    \textbf{Evaluation of estimated depth}:
    It is believed that better performance in synthesizing novel views is directly correlated with the accuracy of depth estimation~\cite{deng2022dsnerf}.
    Thus, we compare our model with DDP-NeRF on their ability to estimate absolute depth correctly using root mean squared error (RMSE).
    We also evaluate the models on their ability to estimate the relative depth of the scene correctly using spearman rank-order correlation coefficient (SROCC)~\cite{corder2014nonparametric}, which computes the linear correlation between ranks of the estimated pixel depths with that of the ground truth depth.
    Due to the unavailability of ground truth depth on both the datasets, we train a NeRF model with dense input views and use its predicted depth as a pseudo ground truth.
    From \cref{tab:quantitative-depth}, we observe that our model consistently outperforms DDP-NeRF both in terms of absolute and relative depth.
    \cref{fig:qualitative-depth} shows that the depth estimated by DDP-NeRF is smooth in textured regions, which may be leading to blur in the synthesized frame.
    In contrast, the dense visibility prior used in our model allows NeRF to predict sharp depth in such regions leading to sharper frame predictions.
    \vspace{1mm}

    \textbf{Ablations:}
    We analyze the contributions of dense visibility and sparse depth priors in ViP-NeRF, by disabling them one at a time.
    From \cref{tab:quantitative-ablations01} and \cref{fig:qualitative-ablations01}, we find that removing either priors leads to a drop in performance on both the datasets.
    This suggests that the dense visibility prior may be providing information that is complementary to the sparse depth prior.
    For a more fine-grained analysis, we compare the LPIPS scores on individual scenes in \cref{fig:quantitative-ablations01}.
    We observe that the addition of dense visibility prior over sparse depth prior leads to an improvement in the performance on all the scenes.
    Further, we find that our model with dense visibility prior alone is able to achieve impressive performance, especially on the RealEstate-10K dataset.

    \subsection{Limitations and Future Work}\label{subsec:limitations}
    Our visibility prior constrains only the regions visible in at least two of the input views.
    As a result, we observe inaccurate depth estimation in the regions that are visible in only one of the input images.
    However, such regions account for a very small portion of the scene and reduce further with three or four input views.
    Further, ViP-NeRF may fail to synthesize disoccluded regions that can occur in sparse-input view synthesis, similar to RegNeRF\@.
    It would be interesting to explore the use of generative NeRF models such as pix2NeRF~\cite{cai2022pix2nerf} to synthesize such disocclusions.

    Our approach to estimating the visibility prior may not account for significant color changes that can occur when the scene contains highly specular surfaces.
    We do not impose any loss on such pixels.
    It would be interesting to analyze if pre-training a network on a large dataset to estimate visibility can provide more supervision in specular regions.
    Moreover, it would be interesting to see if pre-training a network to predict dense visibility generalizes better when compared to depth completion.
    Also, we observe in \cref{tab:quantitative-llff} that adding a new view leads to a significant improvement in performance as compared to adding new regularizations.
    Thus, one could explore hallucinating new views using generative models and use the hallucinated views for additional supervision.

    \section{Conclusion}\label{sec:conclusion}
    We study the problem of training NeRFs in sparse input scenarios, where the NeRF tends to overfit the input views and learn incorrect geometry.
    We propose a prior on the visibility of pixels in other viewpoints to regularize the training and mitigate such errors.
    The visibility prior obtained using a plane sweep volume is more reliable as compared to the depth prior estimated using pre-trained networks.
    We reformulate the NeRF MLPs to additionally output visibility to compute the visibility prior loss in a time-efficient manner.
    ViP-NeRF achieves state-of-the-art performance on two commonly used datasets for novel view synthesis.

\begin{acks}
 This work was supported in part by a grant from Qualcomm.
 The first author was supported by the Prime Minister’s Research Fellowship awarded by the Ministry of Education, Government of India.
 The authors would also like to thank Suhas Srinath and Nithin Babu for the valuable discussions.
\end{acks}

    \begin{figure*}
        \centering
        \includegraphics[width=\linewidth]{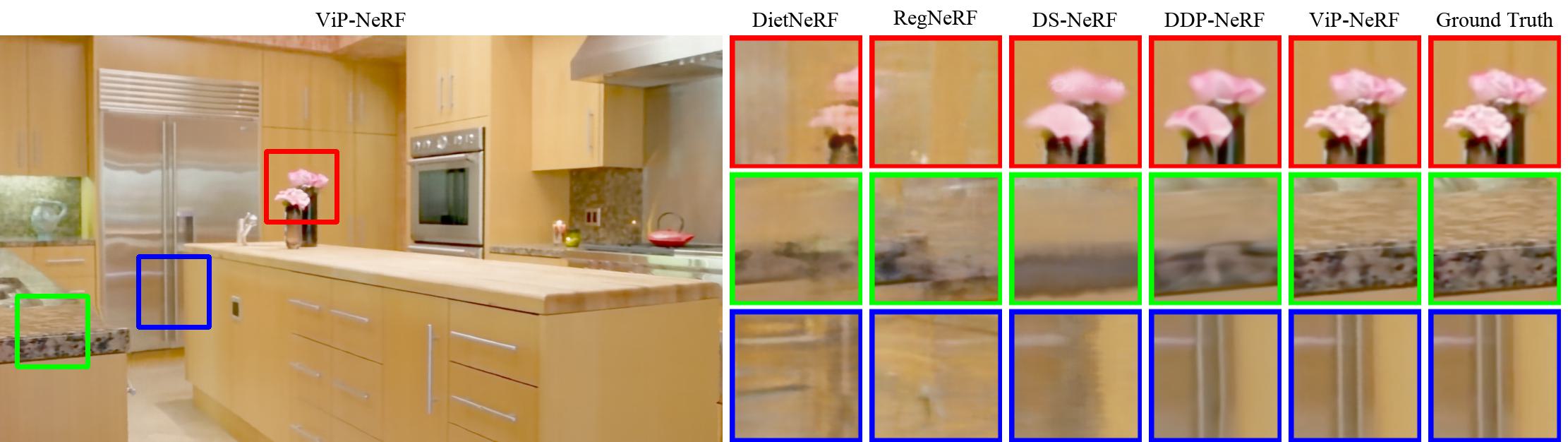}
        \caption{Qualitative examples on RealEstate-10K dataset with two input views.
        We observe sharp predictions by ViP-NeRF while predictions by other models suffer from blur and other artifacts.
        In particular, DDP-NeRF predictions contain blurred flowers (first row) and blurred tiles (second row).
        }
        \label{fig:qualitative-realestate01b}
    \end{figure*}

    \begin{figure*}
        \centering
        \includegraphics[width=\linewidth]{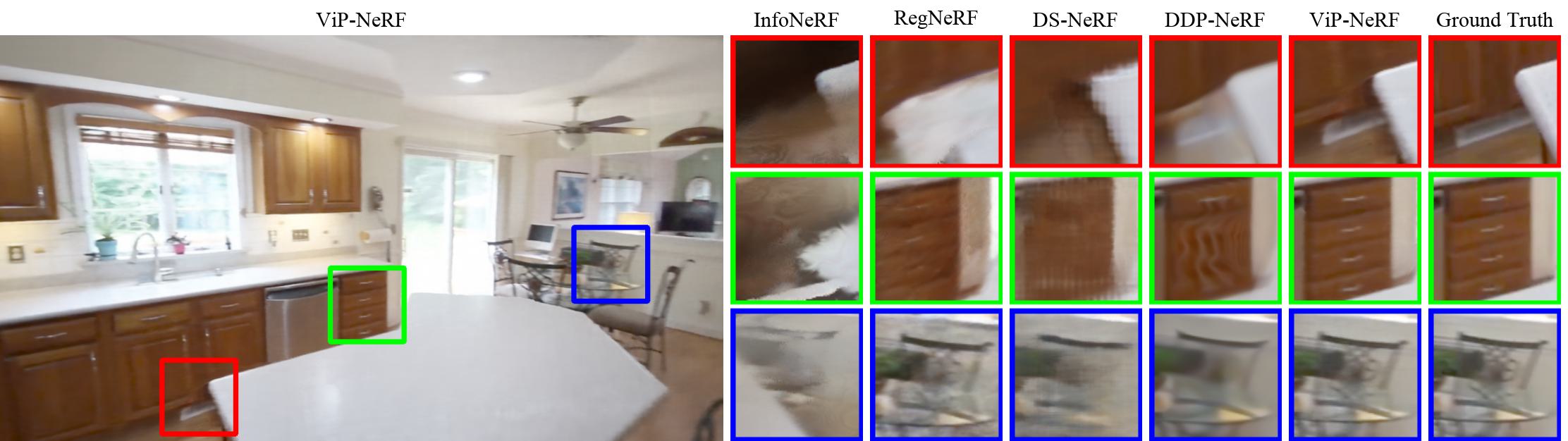}
        \caption{Qualitative examples on RealEstate-10K dataset with three input views.
        We find that ViP-NeRF is able to reconstruct novel views significantly better than the competing models.
        DDP-NeRF extends parts of the white table and fails to reconstruct the drawer handles accurately in the first and second examples.
        In the third example, DDP-NeRF fails to reconstruct thin objects in the chair.
        }
        \label{fig:qualitative-realestate02}
    \end{figure*}

    \begin{figure*}
        \centering
        \includegraphics[width=\linewidth]{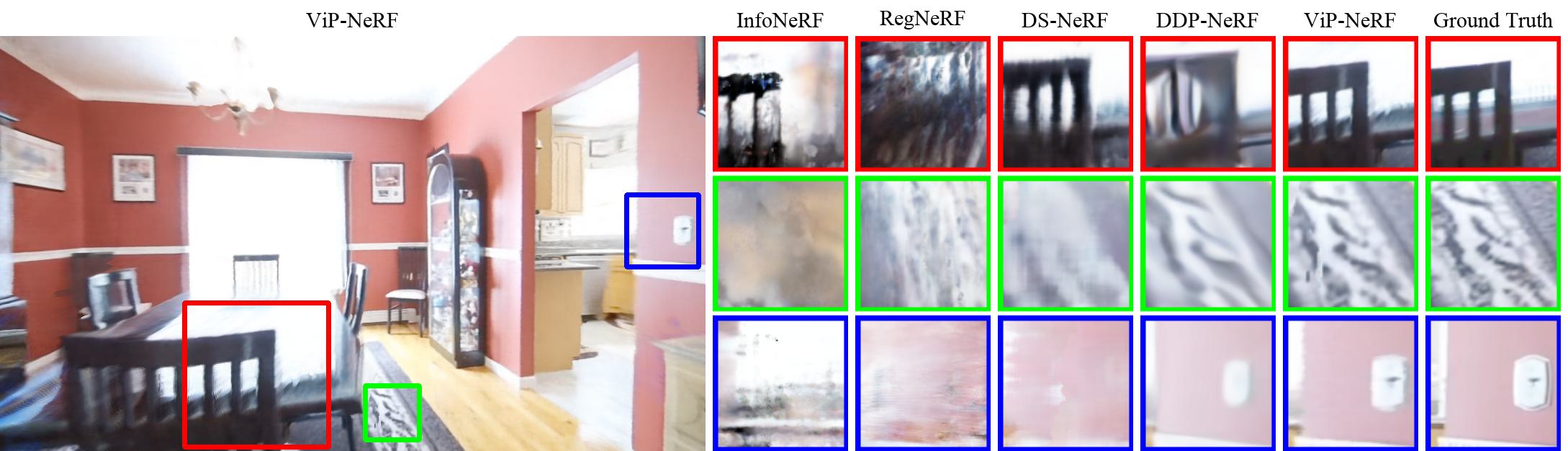}
        \caption{Qualitative examples on RealEstate-10K dataset with four input views.
        In the first example, DDP-NeRF fails to retain the structure of the chair while it blurs the texture of the carpet in the second example.
        We observe even more severe distortions among the predictions of other models.
        }
        \label{fig:qualitative-realestate03}
    \end{figure*}


    \begin{figure*}
        \centering
        \includegraphics[width=\linewidth]{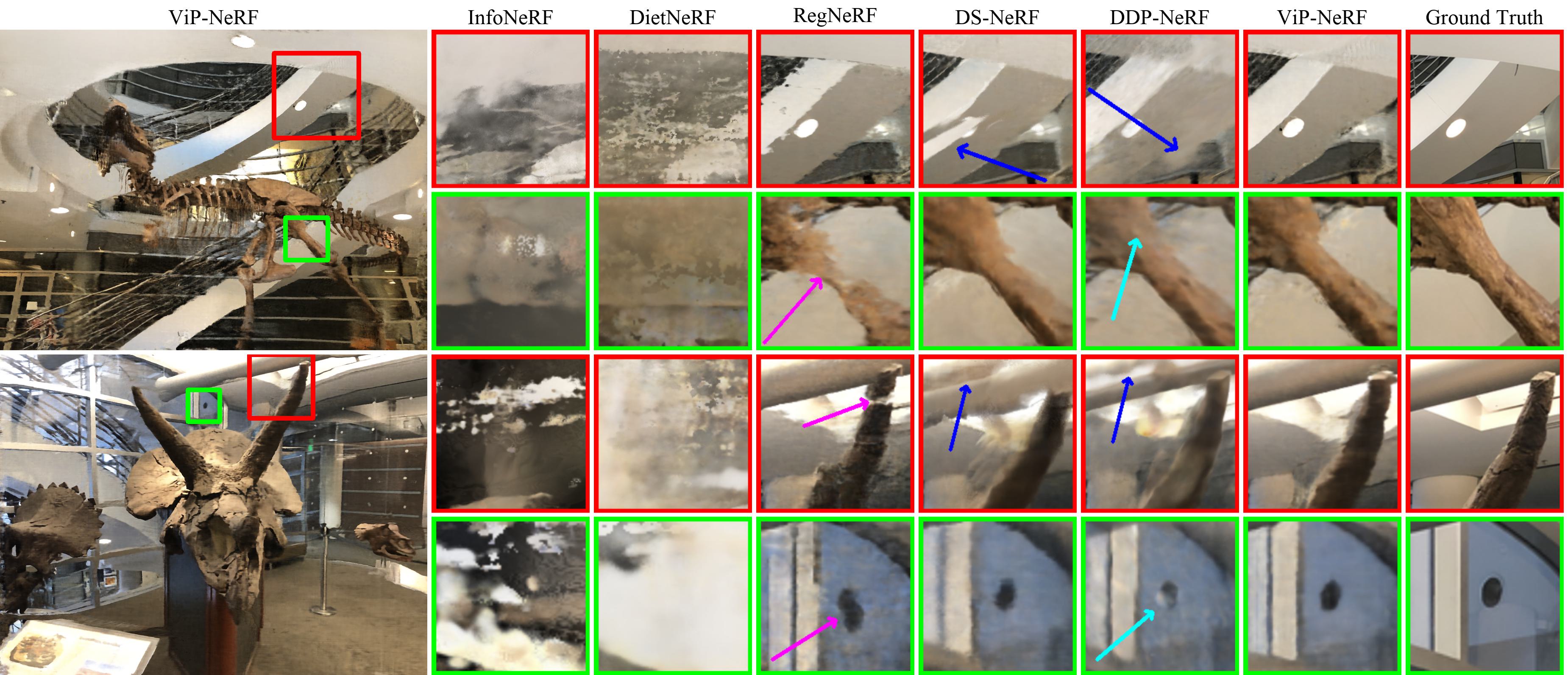}
        \caption{Qualitative examples on NeRF-LLFF dataset with two input views.
        In the first and third examples, we observe floater artifacts (blue arrows) in the predictions of DS-NeRF and DDP-NeRF, which are mitigated in the predictions of ViP-NeRF.
        We find that RegNeRF fails to capture thin t-rex bone in the second example and breaks the horn into two pieces in the third example (magenta arrows). 
        Cyan arrows indicate color changes in the predictions of DDP-NeRF in the second and fourth examples.
        We note that predictions by our model do not suffer from the above described artifacts.
        }
        \label{fig:qualitative-llff01}
    \end{figure*}

    \begin{figure*}
        \centering
        \begin{subfigure}{.48\textwidth}
            \centering
            \includegraphics[width=\linewidth]{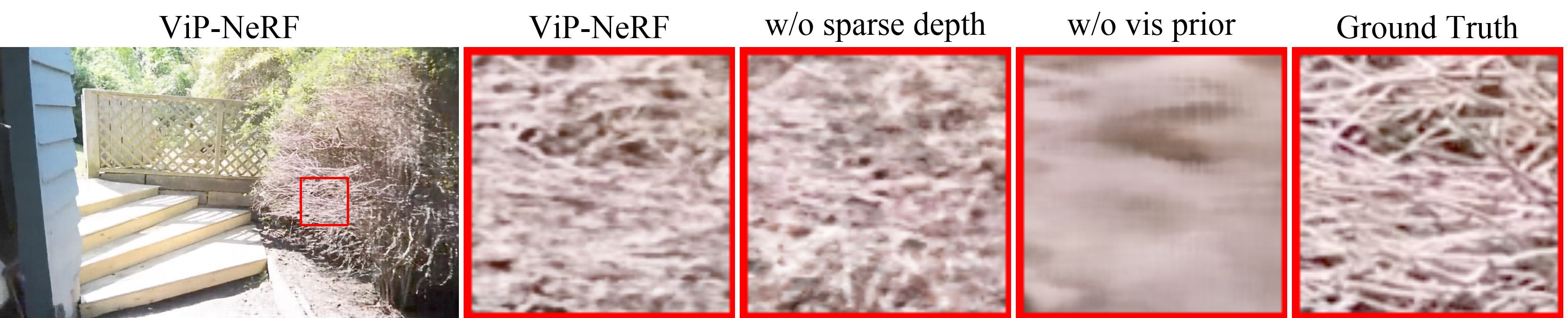}
            \includegraphics[width=\linewidth]{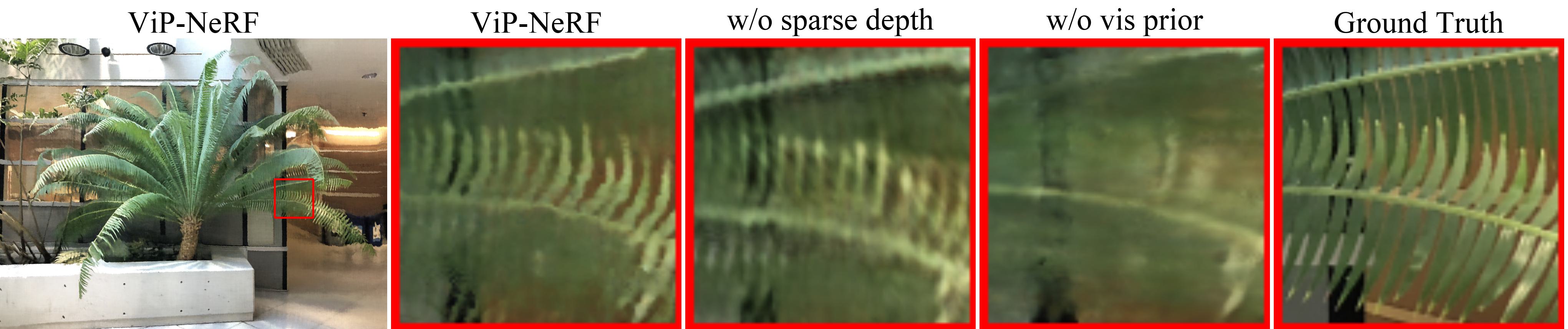}
            \caption{Qualitative examples for ablations on RealEstate-10K and NeRF-LLFF dataset.
            We observe that the absence of dense visibility prior leads to significant blur in the predicted frames.
            While the reconstruction is reasonable without the sparse depth prior, we obtain the best reconstructions when using both the priors.}
            \label{fig:qualitative-ablations01}
        \end{subfigure}\hfill%
        \begin{subfigure}{.48\textwidth}
            \centering
            \begin{subfigure}{0.48\linewidth}
                \centering
                \includegraphics[width=\linewidth]{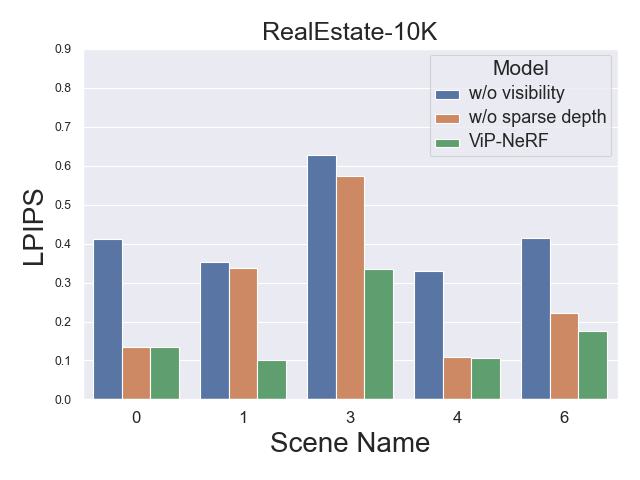}
            \end{subfigure}
            \begin{subfigure}{0.48\linewidth}
                \centering
                \includegraphics[width=\linewidth]{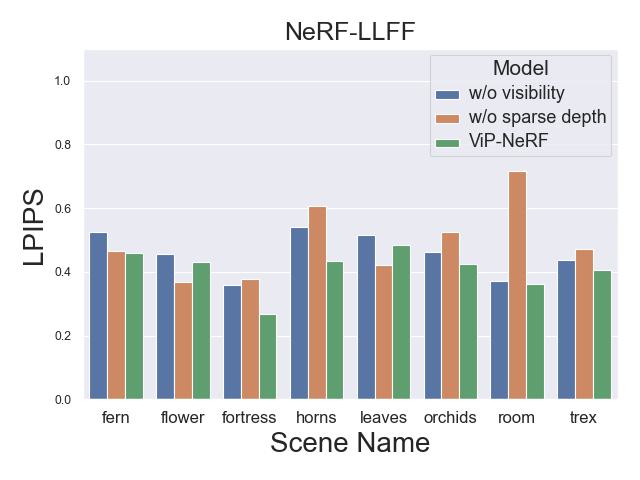}
            \end{subfigure}
            \caption{Scene-wise LPIPS scores of ViP-NeRF and the ablated models.
            Note that lower LPIPS scores are better.
            ViP-NeRF performs better than both the ablated models in most cases leading to overall better performance.}
             \label{fig:quantitative-ablations01}
        \end{subfigure}
        \caption{Qualitative and quantitative comparisons of ablated models on both RealEstate-10K and NeRF-LLFF datasets.}
    \end{figure*}

    \bibliographystyle{ACM-Reference-Format}
    \bibliography{VSTURW}

\appendix
\twocolumn[\subsection*{\centering \fontsize{13}{15}\selectfont Supplement}]

    \noindent The contents of this supplement include
    \begin{enumerate}[label=\Alph*., noitemsep]
        \item Details of RealEstate-10K dataset.
        \item Implementation details of the competing sparse input NeRF models.
        \item Comparisons on DTU dataset.
        \item Video examples on RealEstate-10K and NeRF-LLFF datasets.
        \item Additional comparisons between ViP-NeRF and DDP-NeRF\@.
        \item Additional analysis.
    \end{enumerate}

    \section{Dataset Details}\label{sec:dataset-details}
    \textbf{RealEstate-10K}~\cite{zhou2018stereomag} is a large database consisting of about 80,000 video segments, each containing more than 30 frames.
    This dataset was proposed to train traditional deep learning models which require training on a large number of videos and hence the dataset is further divided into train and test splits.
    Since NeRF based models optimize the networks on individual scenes, we select five videos from the test set to evaluate the NeRF models.
    For easy reference, we rename the videos as scene numbers starting from zero.
    The dataset provides only links to the videos on YouTube and hence we discard the videos which are no longer available.
    Further, we also discard the videos which are less than 50 frames in length.
    We then select the first five videos and choose a random segment of 50 frames within the videos.
    In \cref{tab:realestate-name-mapping}, we provide the mapping between the original name of the videos we selected and the updated names, along with the timestamp of the first frame in the video segment.
    Please refer to ~\citet{zhou2018stereomag} for more details on obtaining the data.

    In each scene, we reserve 5 frames for training and use the remaining 45 frames for testing.
    Specifically, frames 10, 20, 30, 0 and 40 are reserved for training.
    For training with $n$ views, we choose the first $n$ frames from the above list.

    \begin{table}
        \centering
        \caption{Original name of RealEstate-10K videos we selected for experiments and their updated names.}
        \begin{tabular}{c|c|c}
            \hline
            Original Name    & Updated Name & Timestamp \\
            \hline
            000c3ab189999a83 & 0            & 53453400  \\
            000db54a47bd43fe & 1            & 227894333 \\
            0017ce4c6a39d122 & 3            & 46012000  \\
            002ae53df0e0afe2 & 4            & 61144000  \\
            0043978734eec081 & 6            & 54387667  \\
            \hline
        \end{tabular}
        \label{tab:realestate-name-mapping}
    \end{table}

    \section{Implementation Details}\label{sec:implementation-details}
    We use the official code releases by the respective authors and train the models on both RealEstate-10K and NeRF-LLFF datasets~\cite{mildenhall2019llff}.
    In the following, we provide details of any changes we make on top of the respective code releases.

    \textbf{InfoNeRF}~\cite{kim2022infonerf}:
    The code release for InfoNeRF uses test viewpoints during training.
    For a fair comparison, we replace the test poses with poses interpolated from train poses.

    \textbf{RegNeRF}~\cite{niemeyer2022regnerf}:
    We found an inconsistency between the description in the paper and the implementation (possibly a bug) in RegNeRF code, where they use only a single hallucinated view instead of multiple.
    We fix this and train the RegNeRF model.


    \section{Comparisons on DTU Dataset}\label{sec:comparisons-dtu-dataset}
     \begin{table}
         \centering
         \caption{Quantitative results on DTU dataset. RegNeRF+ uses test camera poses during training.}
         \resizebox{\linewidth}{!}{
             \begin{tabular}{l|c|ccc}
                 \hline
                 \textbf{Model} & \textbf{\makecell{learnt\\prior}} & \textbf{LPIPS \textdownarrow} & \textbf{SSIM \textuparrow} & \textbf{PSNR \textuparrow} \\
                 \hline
                 InfoNeRF &  & 0.6649 & 0.2659 & 8.67  \\
                 DietNeRF & \checkmark & 0.7686 & 0.2790 & 7.36  \\
                 RegNeRF &  & 0.7808 & 0.2327 & 7.25  \\
                 \textcolor{gray}{RegNeRF+} &  & \textcolor{gray}{0.4378} & \textcolor{gray}{0.5310} & \textcolor{gray}{12.73}  \\
                 DS-NeRF &  & 0.5136 & 0.4841 & 11.99 \\
                 DDP-NeRF & \checkmark  & 0.5542 & 0.4544 & 11.40 \\
                 ViP-NeRF &  & \textbf{0.4876} & \textbf{0.5057} & \textbf{12.04} \\
                 \hline
             \end{tabular}
         }
         \label{tab:quantitative-dtu}
     \end{table}
    \begin{figure*}
        \centering
        \includegraphics[width=\linewidth]{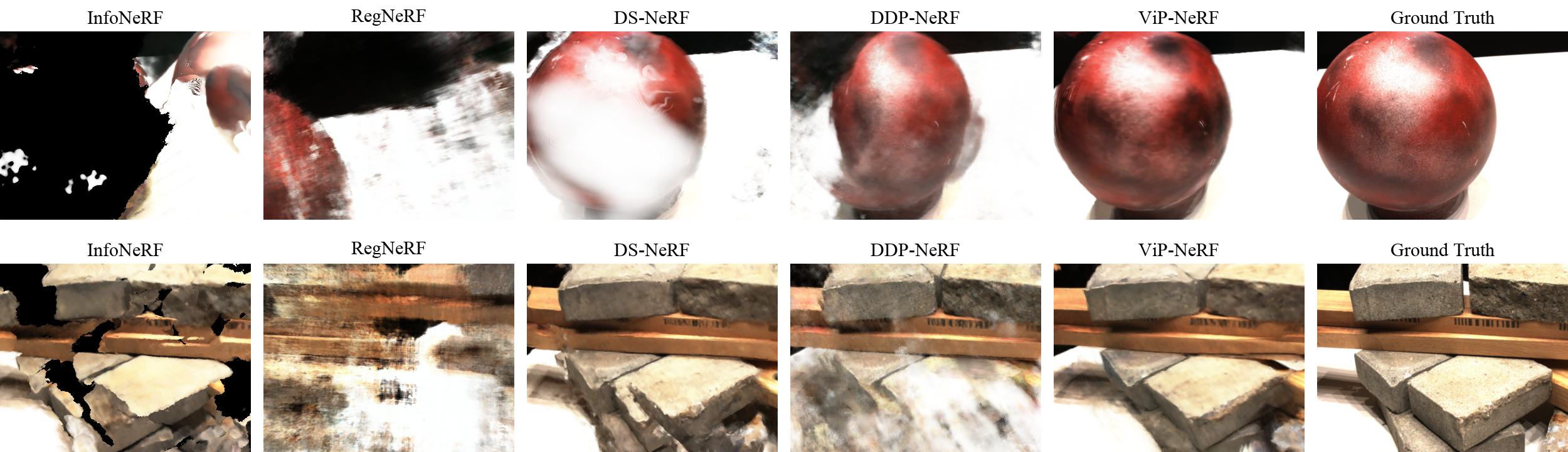}
        \caption{Qualitative examples on DTU dataset. We observe various distortions among the predictions by competing models.
        For example, in the second row, the DS-NeRF prediction has a ghosting artifact in one of the bricks.
        However, predictions by our model are significantly closer to the ground truth.}
        \label{fig:qualitative-dtu01}
    \end{figure*}

    DTU~\cite{aanaes2016dtu} is a commonly used benchmark dataset for conditional NeRF models~\cite{yu2021pixelnerf,chen2021mvsnerf}.
    Nonetheless, a few sparse input NeRF models benchmark on the DTU dataset as well.
    We use the train and test sets defined by pixelNeRF~\cite{yu2021pixelnerf}.
    Specifically, there are 15 scenes and around 49 frames per scene of which 40 frames are used for testing, and 9 frames are reserved for training.

    We found that RegNeRF uses the test viewpoints as the hallucinated viewpoints for training on DTU scenes.
    However, this would be an unfair comparison since other models including ours do not use test camera poses during training.
    Hence we remove such viewpoints and train the RegNeRF model.
    Nonetheless, we report the performance of RegNeRF that uses test views during training as `\textcolor{gray}{RegNeRF+}' in \cref{tab:quantitative-dtu}.

    We show quantitative comparisons between the competing models in \cref{tab:quantitative-dtu} and qualitative comparisons in \cref{fig:qualitative-dtu01}.
    ViP-NeRF outperforms all the competing models that do not use test camera poses during training, including DDP-NeRF and DietNeRF which employ pre-training.
    Even qualitatively, we find that the predictions of ViP-NeRF are closer to the ground truth without many artifacts seen in the predictions of the other models.

    \section{Video Comparisons}\label{sec:video-comparisons}
    Along with this supplementary, we attach a few videos to compare ViP-NeRF with the competing models such as DS-NeRF~\cite{deng2022dsnerf}, DDP-NeRF~\cite{roessle2022ddpnerf} and RegNeRF~\cite{niemeyer2022regnerf}.
    Kindly use the attached `VideoSamples.html' file to view all the videos in a single browser window.

    \section{Additional Comparisons}\label{sec:additional-comparisons}
%

    \begin{figure*}
        \centering
        \includegraphics[width=\linewidth]{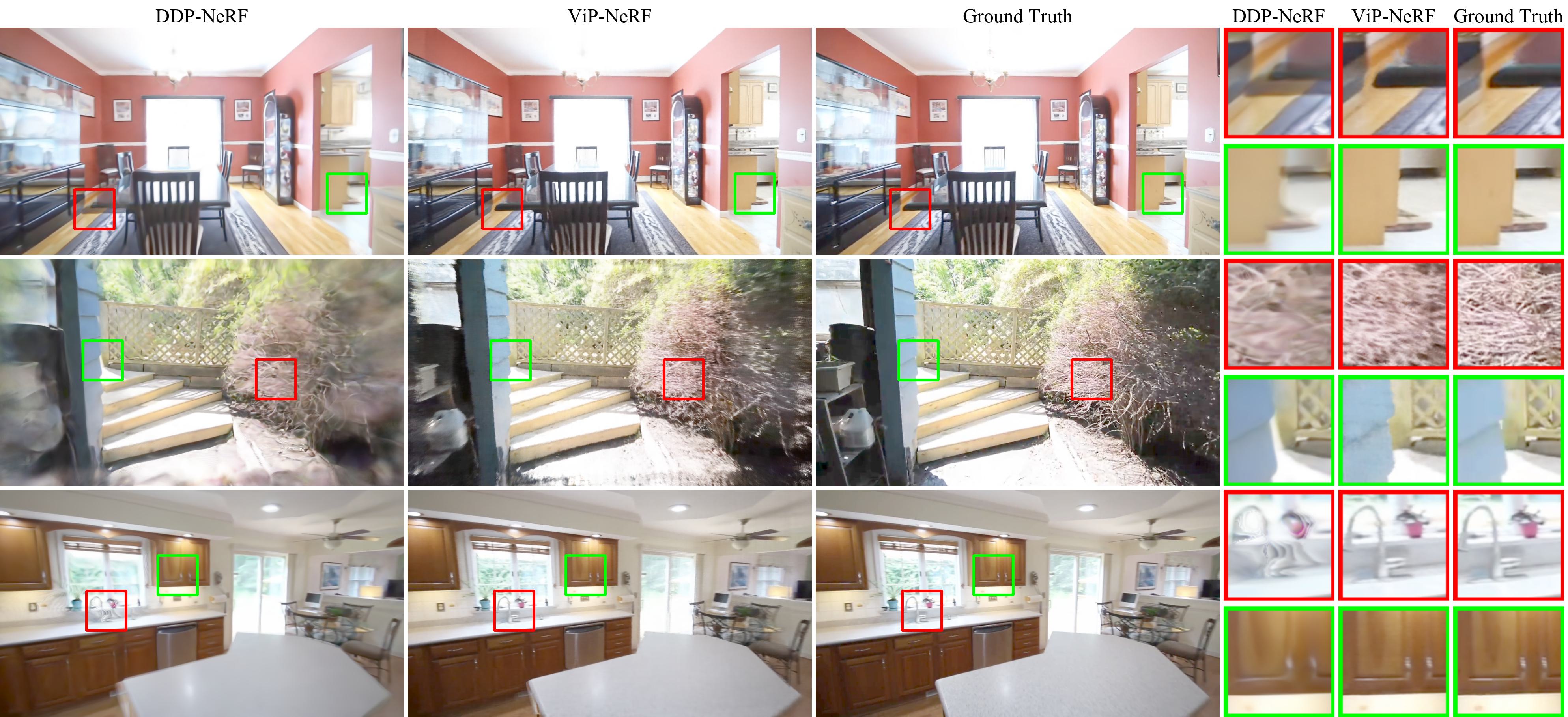}
        \caption{Qualitative examples on RealEstate-10K dataset with two input views.
        In the first example, we see artifacts such as ghosting and blur in the frame predicted by DDP-NeRF.
        In the second example, we observe that DDP-NeRF infers incorrect geometry due to which the objects are placed incorrectly in the synthesized view.
        However, the predictions by ViP-NeRF do not suffer from such artifacts and are significantly sharper.}
        \label{fig:qualitative-realestate11}
    \end{figure*}
    \begin{figure*}
        \centering
        \includegraphics[width=\linewidth]{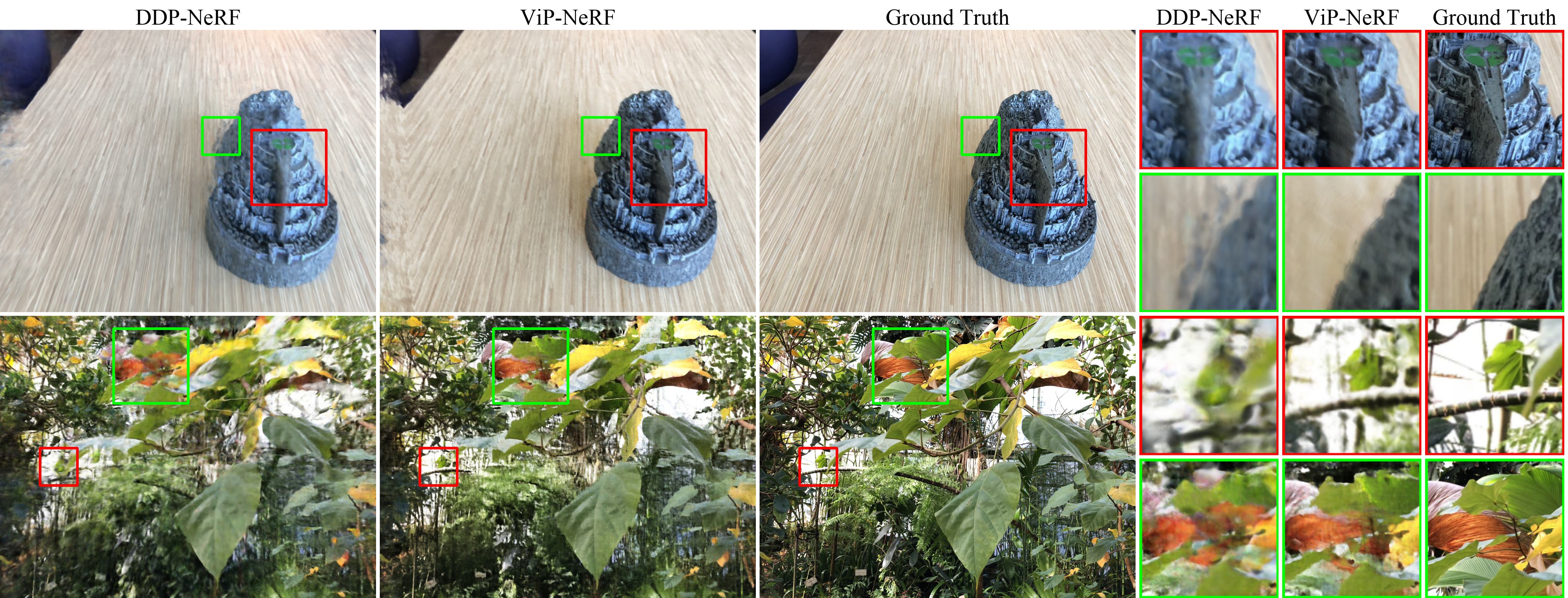}
        \caption{Qualitative examples on NeRF-LLFF dataset with three input views.
        In the first example, we find that DDP-NeRF prediction has a different global color than the ground truth.
        In addition, the angle of the sharp triangular object on the fortress is changed.
        We also observe floating blue clouds outside the fortress and blur in other regions.
        In the second example, we notice that DDP-NeRF is unable to infer the positions of the objects (horizontal stem and the orange leaf) correctly and instead places them at incorrect positions or breaks them into multiple parts.
        On the other hand, ViP-NeRF is able to synthesize the novel views reasonably well.}
        \label{fig:qualitative-llff12}
    \end{figure*}
    \begin{figure*}
        \centering
        \includegraphics[width=\linewidth]{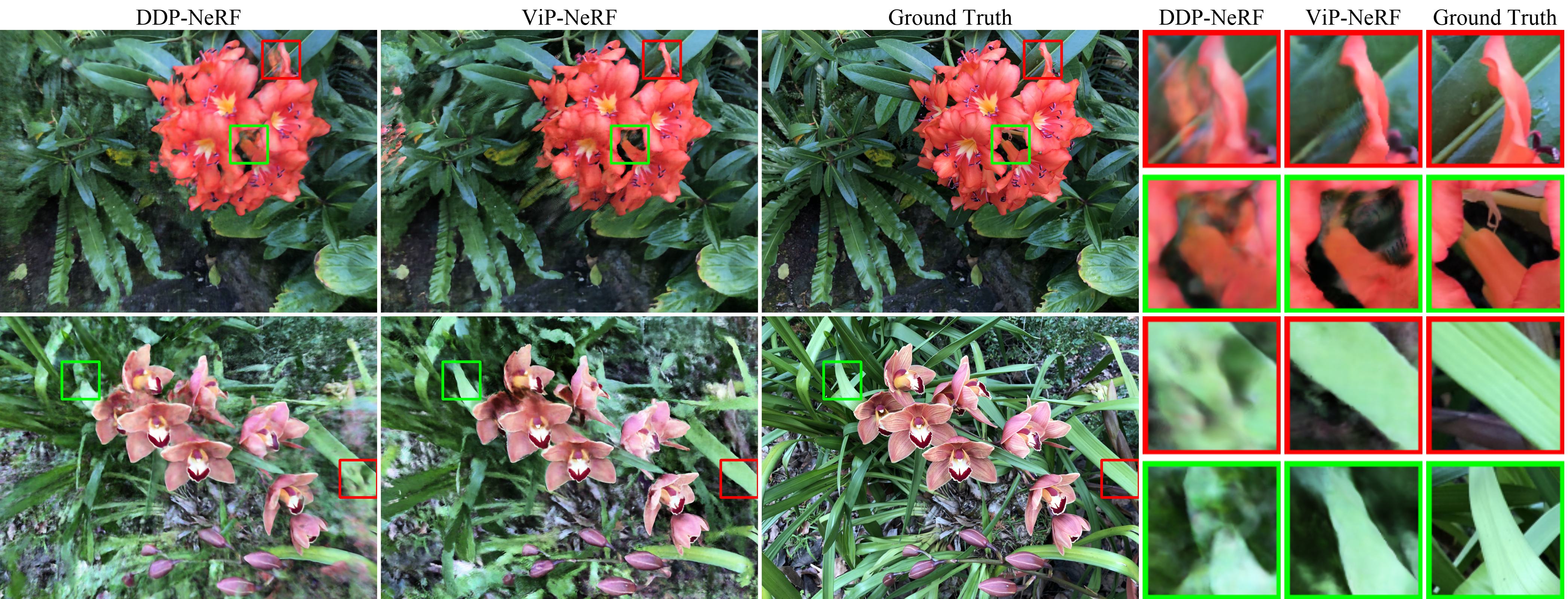}
        \caption{Qualitative examples on NeRF-LLFF dataset with four input views.
        We find that DDP-NeRF predictions contain floaters (orange in the first example and black in the second), whereas ViP-NeRF predictions are devoid of such artifacts.
        We also observe that ViP-NeRF finds difficulty in synthesizing disoccluded regions such as the flower stem in the first example.
        Nonetheless, ViP-NeRF predicts plausible solutions in such cases.}
        \label{fig:qualitative-llff13}
    \end{figure*}
    \begin{figure}
        \centering
        \includegraphics[width=\linewidth]{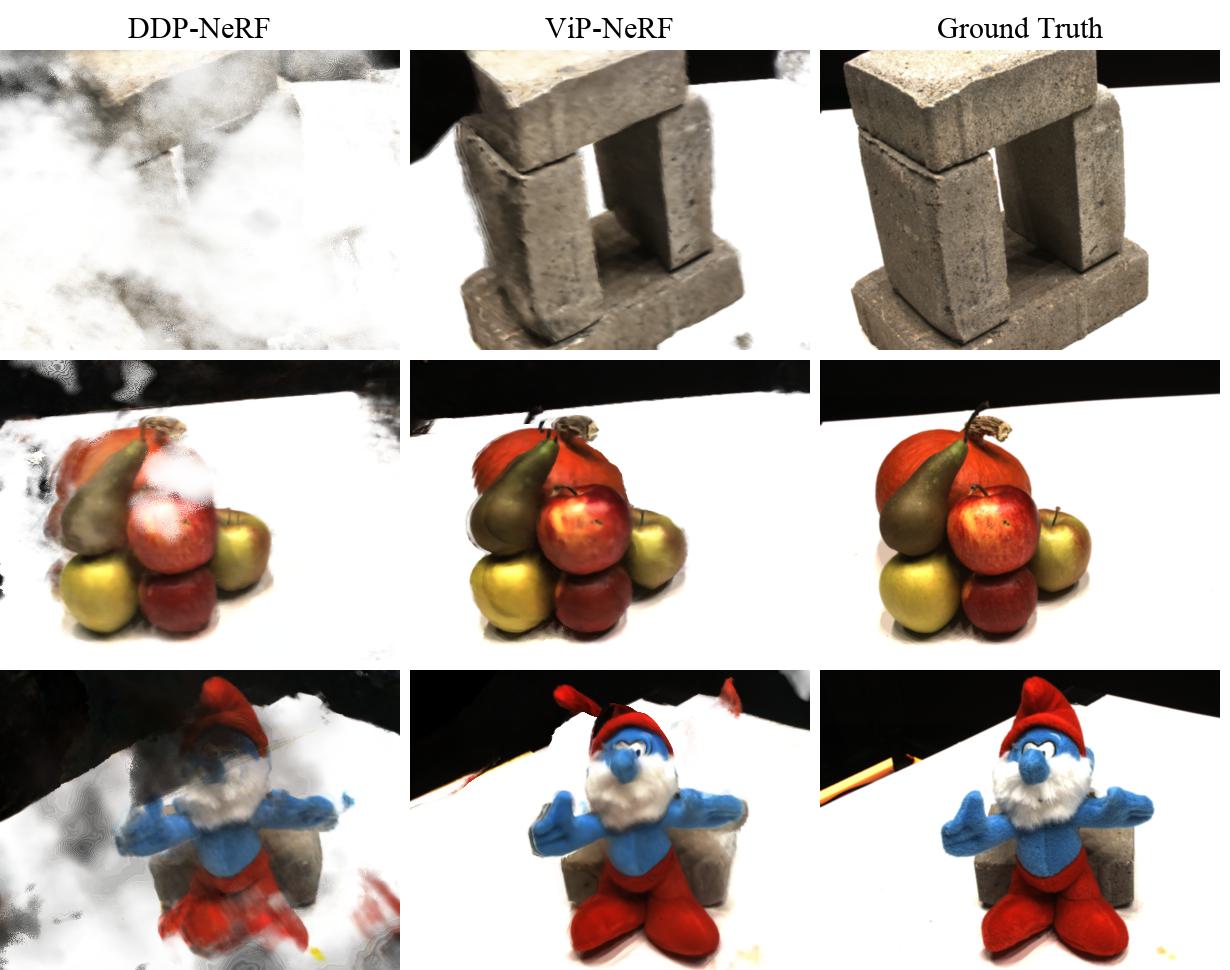}
        \caption{Qualitative examples on DTU dataset with two input views.
        DDP-NeRF predictions contain significant floating clouds in all three examples, whereas ViP-NeRF produces more realistic novel views.}
        \label{fig:qualitative-dtu11}
    \end{figure}

    Here, we show more comparisons with the second-best-performing model, DDP-NeRF, on both datasets.
    Specifically, out of 5 scenes from RealEstate-10K, Figs. 3 and 7 in the main paper show the comparisons on two of the scenes.
    \cref{fig:qualitative-realestate11} shows qualitative comparisons on the remaining three scenes.
    We observe that ViP-NeRF synthesizes superior quality frames compared to DDP-NeRF in all the scenes.
    On the NeRF-LLFF dataset, the figures in the main paper show ViP-NeRF predictions on 4 out of 8 scenes in the dataset.
    \cref{fig:qualitative-llff12,fig:qualitative-llff13} shows ViP-NeRF predictions on the remaining four scenes in the NeRF-LLFF dataset.

    Fig.\ 10 in the main paper shows qualitative comparisons with two input views on the NeRF-LLFF dataset.
    \cref{fig:qualitative-llff12,fig:qualitative-llff13} show qualitative comparisons with three and four input views, respectively.
    From the figures, we observe that ViP-NeRF synthesizes sharper frames as compared to DDP-NeRF while maintaining better structures of the objects.
    Finally, \cref{fig:qualitative-dtu11} shows more comparisons with DDP-NeRF on the DTU dataset.
    We find that while ViP-NeRF synthesizes plausible frames, DDP-NeRF predictions suffer from floating white/black clouds that occlude the objects of interest.

    In \cref{tab:quantitative-scene-wise-realestate01,tab:quantitative-scene-wise-realestate02,tab:quantitative-scene-wise-realestate03,tab:quantitative-scene-wise-llff01,tab:quantitative-scene-wise-llff02,tab:quantitative-scene-wise-llff03}, we provide scene-wise quantitative performance comparisons.

    \section{Additional Analysis}\label{sec:additional-analysis}

    \subsection{PSV based Dense Depth Prior}\label{subsec:psv-dense-depth-prior}
    \begin{table}
        \centering
        \setlength\tabcolsep{2pt}
        \caption{Comparison between NeRF models trained with visibility and depth priors obtained using plane sweep volume.}
        \begin{tabular}{c|cc|cc}
            \hline
            & \multicolumn{2}{c|}{RealEstate-10K} & \multicolumn{2}{c}{NeRF-LLFF} \\
            model                & \textbf{LPIPS \textdownarrow} & \textbf{SSIM \textuparrow} & \textbf{LPIPS \textdownarrow} & \textbf{SSIM \textuparrow} \\
            \hline
            ViP-NeRF             & \textbf{0.1704}               & \textbf{0.8087}            & \textbf{0.4017}               & \textbf{0.5222}            \\  
            DS-NeRF + PSV dense depth   & 0.7453                        & 0.4247                     & 0.6238                        & 0.3878                     \\  
            \hline
        \end{tabular}
        \label{tab:quantitative-ablations02}
    \end{table}
    \begin{figure}
        \centering
        \includegraphics[width=\linewidth]{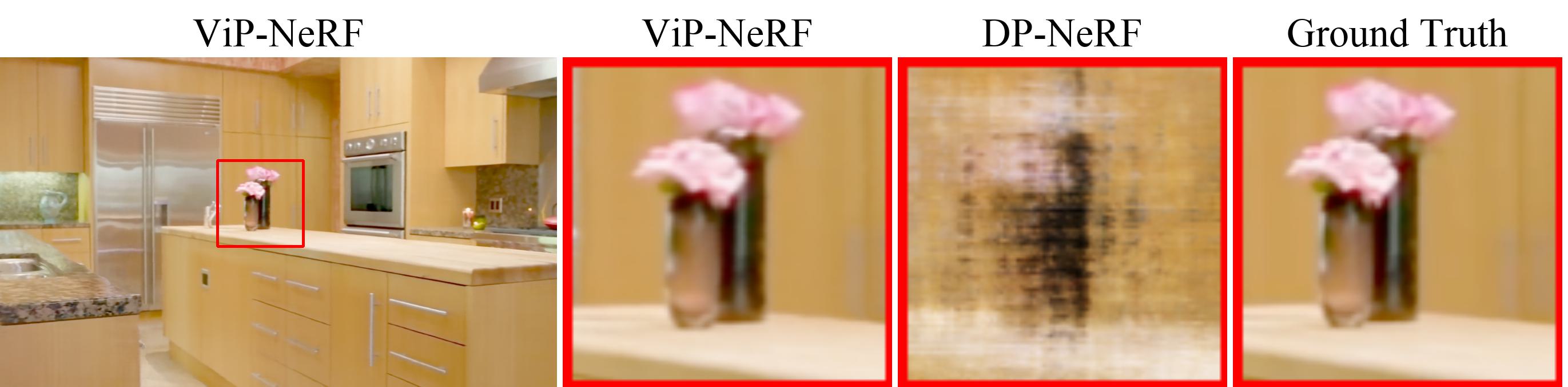}
        \includegraphics[width=\linewidth]{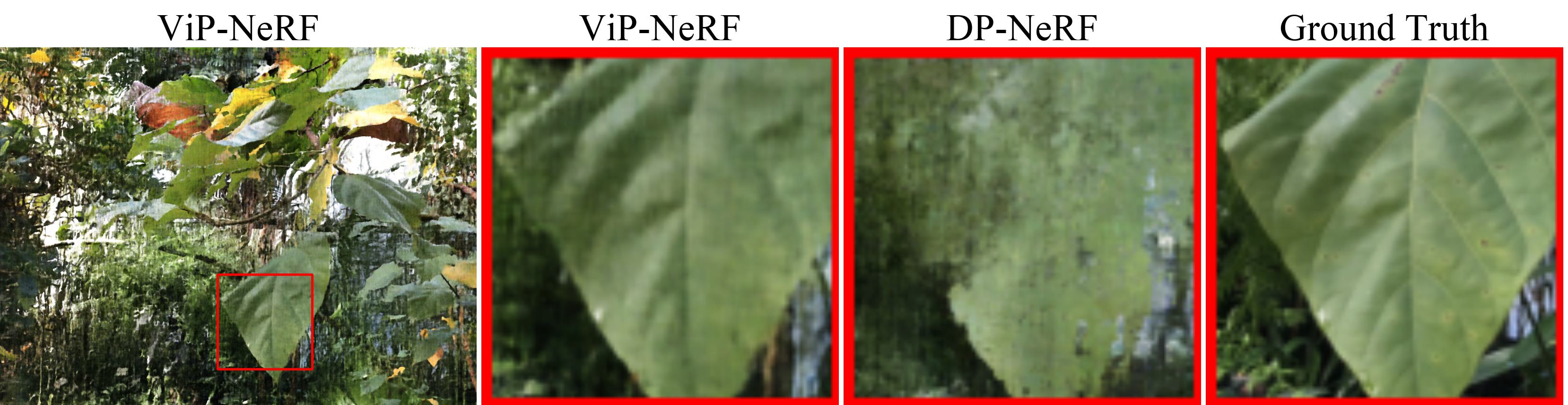}
        \caption{Qualitative examples on RealEstate-10K and NeRF-LLFF datasets with visibility and depth prior (DP-NeRF) obtained using our method.
        We observe significant distortions in the predictions of DP-NeRF due to unreliable depth prior.}
        \label{fig:qualitative-ablations02}
    \end{figure}
    Here, we ask whether the improvement in view synthesis performance is because the visibility prior can be reliably estimated or due to the use of plane sweep volumes.
    To answer this, we experiment with obtaining a dense depth prior using the plane sweep volume similar to our visibility map computation.
    Specifically, we estimate dense depth as $z_p(\mathbf{q}) = \arg \min_k E_k(\mathbf{q})$ similar to Eq.\ 10 and use it to supervise NeRF similar to Eq.\ 13.
    We show the quantitative and qualitative performance of this model in \cref{tab:quantitative-ablations02} and \cref{fig:qualitative-ablations02}, where we find that its performance significantly deteriorates.
    This supports our hypothesis that estimating depth accurately is a very hard problem that may require pre-training on a large dataset, but reliably estimating visibility appears to be relatively easier.

    \subsection{Distant Views}\label{subsec:distant-views}
    \begin{figure*}
        \centering
        \begin{subfigure}{.48\textwidth}
            \centering
            \includegraphics[width=\linewidth]{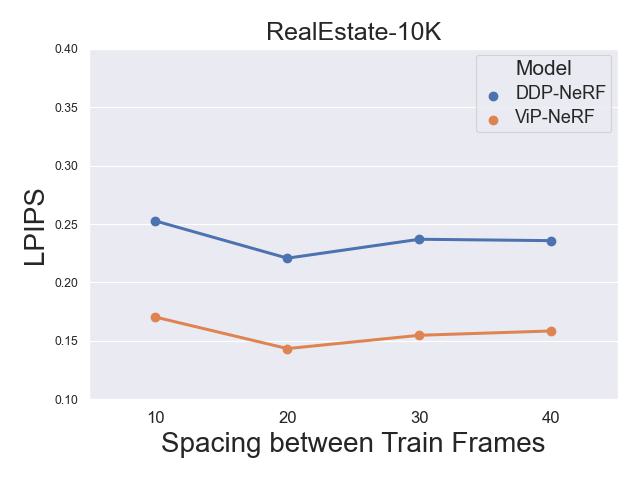}
        \end{subfigure}\hfill%
        \begin{subfigure}{.48\textwidth}
            \centering
            \includegraphics[width=\linewidth]{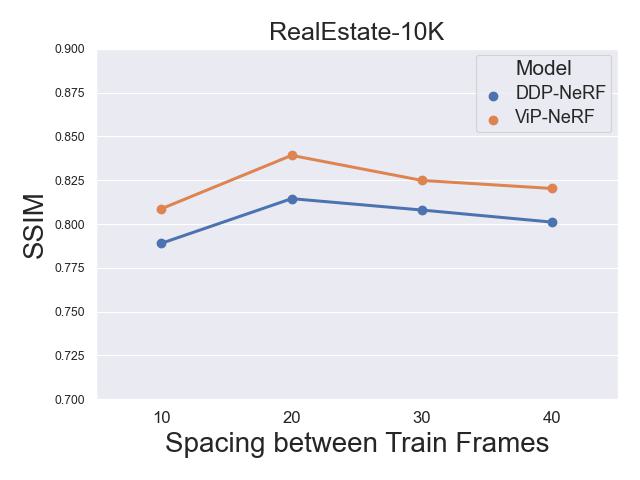}
        \end{subfigure}
        \caption{Quantitative comparison of the performance of DDP-NeRF and ViP-NeRF models with increasing distance between the training views. The x-axis denotes the frames skipped between the two training frames.}
        \label{fig:quantitative-ablations03}
    \end{figure*}
    To understand how the performance of ViP-NeRF varies when the training images are farther apart with more occlusions, we train ViP-NeRF with 2 input views that are 10, 20, 30, and 40 frames apart on the RealEstate-10K dataset.
    We show the performance of ViP-NeRF in terms of LPIPS and SSIM in \cref{fig:quantitative-ablations03}.
    For reference, we also conduct a similar experiment with DDP-NeRF and report its performance.
    We note that the performance of ViP-NeRF stays relatively stable across different settings.
    We observe a small improvement in performance initially, which may be due to the availability of new regions during training when the train views are spread more apart.
    However, the performance tends to drop in small amounts with a further increase in distance between the training views.
    Further, ViP-NeRF outperforms DDP-NeRF in all cases.

    \subsection{Accuracy of Predicted Visibility}\label{subsec:predicted-visibility-accuracy}
    \begin{figure*}
        \centering
        \includegraphics[width=\linewidth]{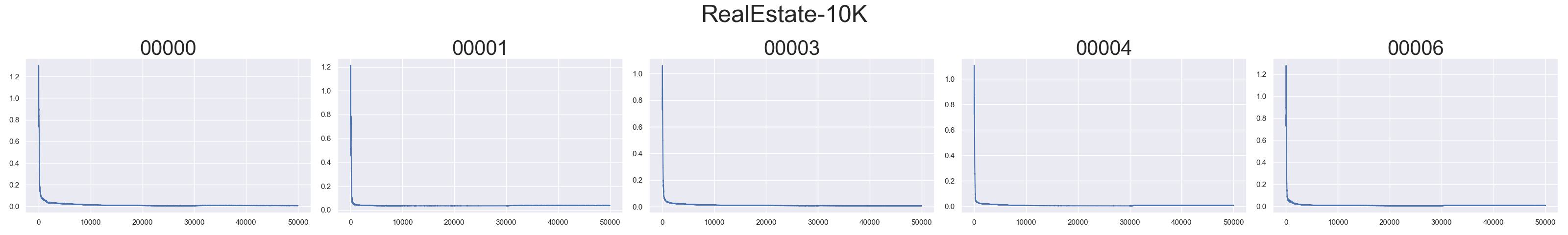}
        \includegraphics[width=\linewidth]{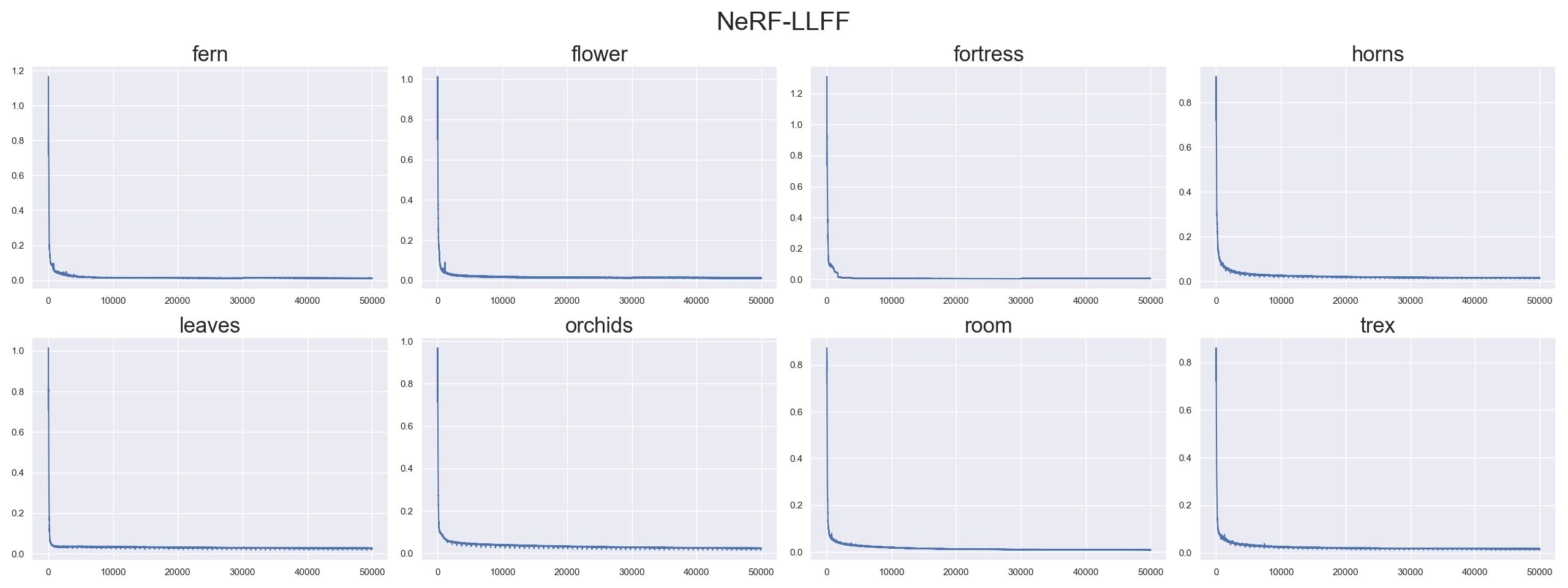}
        \caption{Loss curves of $\mathcal{L}_{v}$ during training on individual scenes of both RealEstate-10K and NeRF-LLFF datasets.}
        \label{fig:visibility-loss-plots}
    \end{figure*}
    Recall that we use Eq.\ 12 during training to enforce the visibility $\hat{T}_i$ output by $\mathcal{F}_2$ to be consistent with the visibility $T_i$ computed by volume rendering.
    In \cref{fig:visibility-loss-plots}, we plot $\mathcal{L}_{v}$ against the iteration number as the training progresses.
    We observe that the loss uniformly decreases to a very small value close to zero indicating that $\hat{T}_i$ and $T_i$ are indeed close to each other.
    This helps effective imposition of our visibility prior.

    \section{Performance on Individual Scenes}\label{sec:performance-on-individual-scenes}
    In \cref{tab:quantitative-scene-wise-realestate01,tab:quantitative-scene-wise-realestate02,tab:quantitative-scene-wise-realestate03,tab:quantitative-scene-wise-llff01,tab:quantitative-scene-wise-llff02,tab:quantitative-scene-wise-llff03}, we report the performance of various models on individual scenes.

    \clearpage
    \begin{table*}
        \centering
        \caption{Per-scene performance of various models with two input views on RealEstate-10K dataset.
        The three rows show LPIPS, SSIM, and PSNR scores, respectively.}
        \begin{tabular}{l|c|c|c|c|c|c}
            \hline
            \textbf{model \textbackslash \ scene name} & \textbf{0} & \textbf{1} & \textbf{3} & \textbf{4}  & \textbf{6}  & \textbf{average} \\
            \hline
            InfoNeRF & \makecell{0.6545 \\0.4863 \\11.85} & \makecell{0.5037 \\0.7458 \\16.19} & \makecell{0.7386 \\0.3002 \\11.59} & \makecell{0.7982 \\0.3990 \\10.55} & \makecell{0.7033 \\0.3953 \\11.32} & \makecell{0.6796 \\0.4653 \\12.30} \\
            \hline
            DietNeRF & \makecell{0.6242 \\0.5743 \\13.16} & \makecell{0.4765 \\0.8298 \\18.13} & \makecell{0.7154 \\0.3065 \\12.11} & \makecell{0.5885 \\0.6632 \\16.86} & \makecell{0.4604 \\0.6919 \\19.22} & \makecell{0.5730 \\0.6131 \\15.90} \\
            \hline
            RegNeRF & \makecell{0.5212 \\0.5416 \\14.83} & \makecell{0.4305 \\0.8163 \\19.11} & \makecell{0.6050 \\0.2549 \\12.88} & \makecell{0.5427 \\0.6797 \\17.78} & \makecell{0.5542 \\0.5618 \\16.13} & \makecell{0.5307 \\0.5709 \\16.14} \\
            \hline
            DS-NeRF & \makecell{0.4128 \\0.7128 \\18.99} & \makecell{0.3535 \\0.8715 \\23.35} & \makecell{0.6269 \\0.4210 \\\textbf{17.08}} & \makecell{0.3298 \\0.8354 \\25.48} & \makecell{0.4133 \\0.7705 \\22.10} & \makecell{0.4273 \\0.7223 \\21.40} \\
            \hline
            DDP-NeRF & \makecell{0.2265 \\0.7985 \\20.79} & \makecell{0.2081 \\0.9222 \\22.16} & \makecell{0.4788 \\\textbf{0.4663} \\16.89} & \makecell{0.1350 \\0.8964 \\24.43} & \makecell{0.2150 \\\textbf{0.8615} \\22.91} & \makecell{0.2527 \\0.7890 \\21.44} \\
            \hline
            ViP-NeRF w/o sparse depth & \makecell{0.1358 \\\textbf{0.8807} \\\textbf{25.54}} & \makecell{0.3363 \\0.8456 \\19.89} & \makecell{0.5723 \\0.3422 \\13.26} & \makecell{0.1100 \\0.9070 \\\textbf{27.98}} & \makecell{0.2225 \\0.8186 \\24.48} & \makecell{0.2754 \\0.7588 \\22.23} \\
            \hline
            ViP-NeRF & \makecell{\textbf{0.1351} \\0.8802 \\25.03} & \makecell{\textbf{0.1008} \\\textbf{0.9425} \\\textbf{26.81}} & \makecell{\textbf{0.3348} \\0.4567 \\17.05} & \makecell{\textbf{0.1053} \\\textbf{0.9085} \\27.91} & \makecell{\textbf{0.1762} \\0.8558 \\\textbf{25.60}} & \makecell{\textbf{0.1704} \\\textbf{0.8087} \\\textbf{24.48}} \\
            \hline
        \end{tabular}
        \label{tab:quantitative-scene-wise-realestate01}
    \end{table*}

    \begin{table*}
        \centering
        \caption{Per-scene performance of various models with three input views on RealEstate-10K dataset.
        The three rows show LPIPS, SSIM, and PSNR scores, respectively.}
        \begin{tabular}{l|c|c|c|c|c|c}
            \hline
            \textbf{model \textbackslash \ scene name} & \textbf{0} & \textbf{1} & \textbf{3} & \textbf{4}  & \textbf{6}  & \textbf{average} \\
            \hline
            InfoNeRF & \makecell{0.7226 \\0.3806 \\10.07} & \makecell{0.6075 \\0.5669 \\13.06} & \makecell{0.6846 \\0.2155 \\10.09} & \makecell{0.7916 \\0.3991 \\10.82} & \makecell{0.6831 \\0.4496 \\11.70} & \makecell{0.6979 \\0.4024 \\11.15} \\
            \hline
            DietNeRF & \makecell{0.5565 \\0.5841 \\14.86} & \makecell{0.5400 \\0.7861 \\16.83} & \makecell{0.6475 \\0.3068 \\12.56} & \makecell{0.5258 \\0.7022 \\18.11} & \makecell{0.4126 \\0.7156 \\20.65} & \makecell{0.5365 \\0.6190 \\16.60} \\
            \hline
            RegNeRF & \makecell{0.4885 \\0.5625 \\15.16} & \makecell{0.3832 \\0.8200 \\20.11} & \makecell{0.5947 \\0.2852 \\13.41} & \makecell{0.5123 \\0.6851 \\18.57} & \makecell{0.3585 \\0.6951 \\19.67} & \makecell{0.4675 \\0.6096 \\17.38} \\
            \hline
            DS-NeRF & \makecell{0.3302 \\0.7772 \\22.09} & \makecell{0.3241 \\0.8902 \\26.61} & \makecell{0.5925 \\0.4540 \\18.30} & \makecell{0.3245 \\0.8460 \\26.94} & \makecell{0.3939 \\0.8096 \\24.70} & \makecell{0.3930 \\0.7554 \\23.73} \\
            \hline
            DDP-NeRF & \makecell{0.1868 \\0.8475 \\22.17} & \makecell{0.1644 \\0.9406 \\23.53} & \makecell{0.4665 \\0.5229 \\17.94} & \makecell{0.1067 \\0.9159 \\27.46} & \makecell{0.1955 \\0.8848 \\24.39} & \makecell{0.2240 \\0.8223 \\23.10} \\
            \hline
            ViP-NeRF w/o sparse depth & \makecell{0.1022 \\0.9046 \\\textbf{27.64}} & \makecell{0.3194 \\0.8284 \\21.04} & \makecell{0.3119 \\\textbf{0.5611} \\\textbf{18.77}} & \makecell{0.0720 \\0.9277 \\31.61} & \makecell{0.1252 \\0.9157 \\29.39} & \makecell{0.1861 \\0.8275 \\25.69} \\
            \hline
            ViP-NeRF & \makecell{\textbf{0.0977} \\\textbf{0.9094} \\27.59} & \makecell{\textbf{0.1302} \\\textbf{0.9408} \\\textbf{28.31}} & \makecell{\textbf{0.3083} \\0.5515 \\18.59} & \makecell{\textbf{0.0682} \\\textbf{0.9319} \\\textbf{32.15}} & \makecell{\textbf{0.1163} \\\textbf{0.9190} \\\textbf{29.40}} & \makecell{\textbf{0.1441} \\\textbf{0.8505} \\\textbf{27.21}} \\
            \hline
        \end{tabular}
        \label{tab:quantitative-scene-wise-realestate02}
    \end{table*}

    \begin{table*}
        \centering
        \caption{Per-scene performance of various models with four input views on RealEstate-10K dataset.
        The three rows show LPIPS, SSIM, and PSNR scores, respectively.}
        \begin{tabular}{l|c|c|c|c|c|c}
            \hline
            \textbf{model \textbackslash \ scene name} & \textbf{0} & \textbf{1} & \textbf{3} & \textbf{4}  & \textbf{6}  & \textbf{average} \\
            \hline
            InfoNeRF & \makecell{0.6386 \\0.5111 \\12.46} & \makecell{0.6315 \\0.4900 \\10.99} & \makecell{0.7153 \\0.2117 \\9.72} & \makecell{0.7993 \\0.3767 \\10.13} & \makecell{0.5879 \\0.5596 \\14.28} & \makecell{0.6745 \\0.4298 \\11.52} \\
            \hline
            DietNeRF & \makecell{0.5724 \\0.5974 \\14.45} & \makecell{0.4908 \\0.8224 \\18.75} & \makecell{0.6502 \\0.3031 \\12.89} & \makecell{0.5386 \\0.6964 \\17.85} & \makecell{0.4164 \\0.7219 \\20.50} & \makecell{0.5337 \\0.6282 \\16.89} \\
            \hline
            RegNeRF & \makecell{0.5054 \\0.5577 \\15.32} & \makecell{0.3902 \\0.8200 \\20.27} & \makecell{0.6256 \\0.2834 \\13.54} & \makecell{0.5229 \\0.6775 \\18.29} & \makecell{0.3711 \\0.6952 \\19.87} & \makecell{0.4831 \\0.6068 \\17.46} \\
            \hline
            DS-NeRF & \makecell{0.3657 \\0.7671 \\22.54} & \makecell{0.3161 \\0.8971 \\27.41} & \makecell{0.6193 \\0.4562 \\18.65} & \makecell{0.3039 \\0.8517 \\27.44} & \makecell{0.3755 \\0.8152 \\25.17} & \makecell{0.3961 \\0.7575 \\24.24} \\
            \hline
            DDP-NeRF & \makecell{0.1834 \\0.8479 \\22.26} & \makecell{0.1307 \\\textbf{0.9490} \\26.08} & \makecell{0.4746 \\0.5394 \\18.65} & \makecell{0.1049 \\0.9161 \\28.81} & \makecell{0.2015 \\0.8828 \\25.04} & \makecell{0.2190 \\0.8270 \\24.17} \\
            \hline
            ViP-NeRF w/o sparse depth & \makecell{0.1022 \\0.9046 \\\textbf{27.64}} & \makecell{0.3194 \\0.8284 \\21.04} & \makecell{0.3119 \\\textbf{0.5611} \\\textbf{18.77}} & \makecell{0.0720 \\0.9277 \\31.61} & \makecell{0.1252 \\0.9157 \\29.39} & \makecell{0.1861 \\0.8275 \\25.69} \\
            \hline
            ViP-NeRF & \makecell{\textbf{0.0977} \\\textbf{0.9094} \\27.59} & \makecell{\textbf{0.1302} \\0.9408 \\\textbf{28.31}} & \makecell{\textbf{0.3083} \\0.5515 \\18.59} & \makecell{\textbf{0.0682} \\\textbf{0.9319} \\\textbf{32.15}} & \makecell{\textbf{0.1163} \\\textbf{0.9190} \\\textbf{29.40}} & \makecell{\textbf{0.1441} \\\textbf{0.8505} \\\textbf{27.21}} \\
            \hline
        \end{tabular}
        \label{tab:quantitative-scene-wise-realestate03}
    \end{table*}

    \begin{table*}
        \centering
        \caption{Per-scene performance of various models with two input views on NeRF-LLFF dataset.
        The three rows show LPIPS, SSIM, and PSNR scores, respectively.}
        \begin{tabular}{l|c|c|c|c|c|c|c|c|c}
            \hline
            \textbf{model \textbackslash \ scene name} & \textbf{fern} & \textbf{flower} & \textbf{fortress} & \textbf{horns} & \textbf{leaves} & \textbf{orchids} & \textbf{room} & \textbf{trex} & \textbf{average} \\
            \hline
            InfoNeRF & \makecell{0.7805 \\0.2635 \\11.00} & \makecell{0.6880 \\0.1890 \\10.97} & \makecell{0.8156 \\0.1808 \\6.48} & \makecell{0.7649 \\0.2020 \\8.86} & \makecell{0.6422 \\0.0987 \\9.47} & \makecell{0.6908 \\0.1108 \\9.43} & \makecell{0.8044 \\0.3757 \\10.76} & \makecell{0.7941 \\0.2115 \\8.44} & \makecell{0.7561 \\0.2095 \\9.23} \\
            \hline
            DietNeRF & \makecell{0.7663 \\0.2883 \\12.30} & \makecell{0.6935 \\0.2590 \\12.18} & \makecell{0.6621 \\0.4424 \\14.22} & \makecell{0.7574 \\0.2819 \\10.71} & \makecell{0.6720 \\0.1196 \\10.58} & \makecell{0.7256 \\0.1470 \\10.57} & \makecell{0.7674 \\0.5153 \\13.09} & \makecell{0.7496 \\0.3674 \\11.30} & \makecell{0.7265 \\0.3209 \\11.89} \\
            \hline
            RegNeRF & \makecell{0.5067 \\0.4681 \\16.51} & \makecell{0.4408 \\\textbf{0.5067} \\\textbf{16.92}} & \makecell{0.3838 \\0.4621 \\20.53} & \makecell{0.5301 \\0.4277 \\15.91} & \makecell{\textbf{0.3590} \\\textbf{0.3637} \\14.51} & \makecell{0.4595 \\0.3018 \\13.88} & \makecell{0.3955 \\0.7306 \\18.59} & \makecell{0.4308 \\\textbf{0.5388} \\\textbf{16.69}} & \makecell{0.4402 \\0.4872 \\16.90} \\
            \hline
            DS-NeRF & \makecell{0.5249 \\0.4681 \\16.69} & \makecell{0.4578 \\0.4455 \\16.17} & \makecell{0.3591 \\0.6316 \\\textbf{23.10}} & \makecell{0.5395 \\0.4816 \\16.64} & \makecell{0.5174 \\0.2429 \\12.68} & \makecell{0.4614 \\0.3190 \\13.86} & \makecell{0.3727 \\\textbf{0.7607} \\\textbf{18.94}} & \makecell{0.4386 \\0.5295 \\15.85} & \makecell{0.4548 \\0.5068 \\17.06} \\
            \hline
            DDP-NeRF & \makecell{0.4715 \\\textbf{0.4940} \\\textbf{17.35}} & \makecell{0.4803 \\0.4535 \\16.18} & \makecell{\textbf{0.1888} \\\textbf{0.7600} \\22.77} & \makecell{0.4973 \\0.5167 \\17.10} & \makecell{0.5550 \\0.2282 \\12.65} & \makecell{0.4438 \\\textbf{0.3682} \\\textbf{15.12}} & \makecell{\textbf{0.3290} \\0.7572 \\18.68} & \makecell{0.4660 \\0.5358 \\15.76} & \makecell{0.4223 \\\textbf{0.5377} \\\textbf{17.21}} \\
            \hline
            ViP-NeRF w/o sparse depth & \makecell{0.4665 \\0.4719 \\16.60} & \makecell{\textbf{0.3673} \\0.5044 \\16.78} & \makecell{0.3779 \\0.5286 \\21.22} & \makecell{0.6068 \\0.4279 \\15.65} & \makecell{0.4210 \\0.3575 \\\textbf{14.76}} & \makecell{0.5252 \\0.2913 \\14.21} & \makecell{0.7174 \\0.5191 \\13.76} & \makecell{0.4706 \\0.5241 \\16.50} & \makecell{0.5056 \\0.4631 \\16.28} \\
            \hline
            ViP-NeRF & \makecell{\textbf{0.4605} \\0.4586 \\16.45} & \makecell{0.4297 \\0.4355 \\14.65} & \makecell{0.2689 \\0.6869 \\22.36} & \makecell{\textbf{0.4356} \\\textbf{0.5353} \\\textbf{17.25}} & \makecell{0.4842 \\0.2172 \\11.90} & \makecell{\textbf{0.4238} \\0.3592 \\14.21} & \makecell{0.3626 \\0.7371 \\18.11} & \makecell{\textbf{0.4052} \\0.5383 \\16.13} & \makecell{\textbf{0.4017} \\0.5222 \\16.76} \\
            \hline
        \end{tabular}
        \label{tab:quantitative-scene-wise-llff01}
    \end{table*}

    \begin{table*}
        \centering
        \caption{Per-scene performance of various models with three input views on NeRF-LLFF dataset.
        The three rows show LPIPS, SSIM, and PSNR scores, respectively.}
        \begin{tabular}{l|c|c|c|c|c|c|c|c|c}
            \hline
            \textbf{model \textbackslash \ scene name} & \textbf{fern} & \textbf{flower} & \textbf{fortress} & \textbf{horns} & \textbf{leaves} & \textbf{orchids} & \textbf{room} & \textbf{trex} & \textbf{average} \\
            \hline
            InfoNeRF & \makecell{0.8583 \\0.2144 \\7.51} & \makecell{0.6949 \\0.2308 \\10.75} & \makecell{0.8272 \\0.1680 \\5.22} & \makecell{0.7555 \\0.1800 \\8.87} & \makecell{0.6518 \\0.1083 \\9.92} & \makecell{0.7121 \\0.0834 \\8.28} & \makecell{0.8312 \\0.2911 \\8.94} & \makecell{0.7885 \\0.1768 \\8.75} & \makecell{0.7679 \\0.1859 \\8.52} \\
            \hline
            DietNeRF & \makecell{0.7977 \\0.3207 \\12.01} & \makecell{0.6705 \\0.3052 \\13.08} & \makecell{0.6984 \\0.4098 \\12.37} & \makecell{0.7458 \\0.2946 \\11.10} & \makecell{0.6672 \\0.1266 \\10.57} & \makecell{0.7478 \\0.1493 \\10.13} & \makecell{0.7620 \\0.5518 \\12.62} & \makecell{0.7223 \\0.3514 \\11.88} & \makecell{0.7254 \\0.3297 \\11.77} \\
            \hline
            RegNeRF & \makecell{\textbf{0.4874} \\0.4834 \\17.84} & \makecell{\textbf{0.2855} \\0.5764 \\19.48} & \makecell{0.3340 \\0.6222 \\22.62} & \makecell{0.4706 \\0.5177 \\18.12} & \makecell{\textbf{0.4040} \\0.3645 \\14.61} & \makecell{0.4540 \\0.3058 \\14.11} & \makecell{0.2782 \\0.8074 \\20.90} & \makecell{0.3685 \\\textbf{0.6214} \\\textbf{18.42}} & \makecell{0.3800 \\0.5600 \\18.62} \\
            \hline
            DS-NeRF & \makecell{0.5146 \\0.5191 \\\textbf{18.64}} & \makecell{0.3064 \\\textbf{0.6383} \\\textbf{21.35}} & \makecell{0.3024 \\0.6972 \\\textbf{24.63}} & \makecell{0.5185 \\0.5131 \\17.59} & \makecell{0.5533 \\0.2484 \\12.85} & \makecell{0.4811 \\0.3289 \\14.15} & \makecell{\textbf{0.2528} \\\textbf{0.8335} \\\textbf{22.92}} & \makecell{0.4057 \\0.5861 \\17.31} & \makecell{0.4077 \\0.5686 \\\textbf{19.02}} \\
            \hline
            DDP-NeRF & \makecell{0.5186 \\\textbf{0.5322} \\18.61} & \makecell{0.3177 \\0.6216 \\20.34} & \makecell{\textbf{0.2216} \\\textbf{0.7459} \\22.50} & \makecell{0.5187 \\0.5252 \\17.43} & \makecell{0.5602 \\0.2396 \\12.84} & \makecell{0.4860 \\\textbf{0.3518} \\\textbf{15.19}} & \makecell{0.3328 \\0.7659 \\18.65} & \makecell{0.4512 \\0.5397 \\16.26} & \makecell{0.4178 \\0.5610 \\17.90} \\
            \hline
            ViP-NeRF w/o sparse depth & \makecell{0.6117 \\0.4362 \\16.31} & \makecell{0.4248 \\0.4961 \\18.15} & \makecell{0.2978 \\0.6455 \\23.66} & \makecell{0.6195 \\0.4585 \\16.34} & \makecell{0.4625 \\\textbf{0.3743} \\\textbf{15.12}} & \makecell{0.5381 \\0.2938 \\14.37} & \makecell{0.4759 \\0.7304 \\19.21} & \makecell{0.4739 \\0.5127 \\16.60} & \makecell{0.4855 \\0.5110 \\17.71} \\
            \hline
            ViP-NeRF & \makecell{0.5529 \\0.4958 \\17.49} & \makecell{0.2888 \\0.6344 \\20.82} & \makecell{0.2354 \\0.7412 \\24.12} & \makecell{\textbf{0.4632} \\\textbf{0.5619} \\\textbf{18.27}} & \makecell{0.4842 \\0.2522 \\12.61} & \makecell{\textbf{0.4416} \\0.3441 \\14.24} & \makecell{0.2937 \\0.8116 \\21.97} & \makecell{\textbf{0.3483} \\0.6061 \\18.16} & \makecell{\textbf{0.3750} \\\textbf{0.5837} \\18.92} \\
            \hline
        \end{tabular}
        \label{tab:quantitative-scene-wise-llff02}
    \end{table*}

    \begin{table*}
        \centering
        \caption{Per-scene performance of various models with four input views on NeRF-LLFF dataset.
        The three rows show LPIPS, SSIM, and PSNR scores, respectively.}
        \begin{tabular}{l|c|c|c|c|c|c|c|c|c}
            \hline
            \textbf{model \textbackslash \ scene name} & \textbf{fern} & \textbf{flower} & \textbf{fortress} & \textbf{horns} & \textbf{leaves} & \textbf{orchids} & \textbf{room} & \textbf{trex} & \textbf{average} \\
            \hline
            InfoNeRF & \makecell{0.7962 \\0.1901 \\9.83} & \makecell{0.6818 \\0.2296 \\11.54} & \makecell{0.8686 \\0.1608 \\4.72} & \makecell{0.7722 \\0.1853 \\8.81} & \makecell{0.6574 \\0.0904 \\9.30} & \makecell{0.7578 \\0.0851 \\8.12} & \makecell{0.7812 \\0.4544 \\11.89} & \makecell{0.7973 \\0.2589 \\10.10} & \makecell{0.7701 \\0.2188 \\9.25} \\
            \hline
            DietNeRF & \makecell{0.8035 \\0.3473 \\12.87} & \makecell{0.6901 \\0.2921 \\12.57} & \makecell{0.7050 \\0.4186 \\12.62} & \makecell{0.7758 \\0.3036 \\10.81} & \makecell{0.6992 \\0.1366 \\10.81} & \makecell{0.7700 \\0.1565 \\10.15} & \makecell{0.7314 \\0.5944 \\13.91} & \makecell{0.7486 \\0.3507 \\11.15} & \makecell{0.7396 \\0.3404 \\11.84} \\
            \hline
            RegNeRF & \makecell{\textbf{0.3825} \\\textbf{0.6221} \\20.87} & \makecell{0.2981 \\\textbf{0.6378} \\19.80} & \makecell{0.3904 \\0.5383 \\22.23} & \makecell{\textbf{0.3772} \\\textbf{0.6230} \\\textbf{20.10}} & \makecell{\textbf{0.3384} \\\textbf{0.4248} \\\textbf{15.93}} & \makecell{0.4463 \\0.3315 \\14.73} & \makecell{0.2105 \\0.8657 \\23.84} & \makecell{\textbf{0.3454} \\\textbf{0.6503} \\\textbf{18.75}} & \makecell{\textbf{0.3446} \\0.6056 \\19.83} \\
            \hline
            DS-NeRF & \makecell{0.3945 \\0.6172 \\\textbf{20.96}} & \makecell{0.3165 \\0.6285 \\\textbf{20.69}} & \makecell{0.3601 \\0.6431 \\\textbf{24.05}} & \makecell{0.4569 \\0.5766 \\19.52} & \makecell{0.4684 \\0.3721 \\15.81} & \makecell{0.4521 \\0.3803 \\15.40} & \makecell{\textbf{0.1948} \\\textbf{0.8794} \\\textbf{25.35}} & \makecell{0.4307 \\0.5884 \\17.31} & \makecell{0.3825 \\0.6016 \\\textbf{20.11}} \\
            \hline
            DDP-NeRF & \makecell{0.4593 \\0.5849 \\19.75} & \makecell{0.3334 \\0.6118 \\19.83} & \makecell{\textbf{0.2080} \\\textbf{0.7136} \\22.99} & \makecell{0.4718 \\0.5695 \\19.00} & \makecell{0.4921 \\0.3533 \\15.02} & \makecell{0.4584 \\\textbf{0.3937} \\\textbf{15.72}} & \makecell{0.2729 \\0.8227 \\21.82} & \makecell{0.4177 \\0.6031 \\17.57} & \makecell{0.3821 \\0.5999 \\19.19} \\
            \hline
            ViP-NeRF w/o sparse depth & \makecell{0.4951 \\0.5754 \\19.16} & \makecell{0.3100 \\0.6116 \\19.49} & \makecell{0.3733 \\0.5998 \\23.19} & \makecell{0.5117 \\0.5470 \\18.20} & \makecell{0.4615 \\0.3778 \\15.68} & \makecell{0.5283 \\0.3106 \\14.65} & \makecell{0.3130 \\0.8194 \\22.00} & \makecell{0.4061 \\0.6215 \\18.65} & \makecell{0.4197 \\0.5763 \\19.15} \\
            \hline
            ViP-NeRF & \makecell{0.4298 \\0.5788 \\19.35} & \makecell{\textbf{0.2970} \\0.6248 \\19.82} & \makecell{0.2970 \\0.6866 \\23.81} & \makecell{0.4324 \\0.5801 \\19.00} & \makecell{0.4316 \\0.3760 \\14.96} & \makecell{\textbf{0.4295} \\0.3869 \\15.13} & \makecell{0.2607 \\0.8402 \\23.19} & \makecell{0.3462 \\0.6363 \\18.62} & \makecell{0.3593 \\\textbf{0.6085} \\19.58} \\
            \hline
        \end{tabular}
        \label{tab:quantitative-scene-wise-llff03}
    \end{table*}

\end{document}